\documentclass[10pt,twocolumn,letterpaper]{article}

\usepackage[pagenumbers]{cvpr}              %

\newcommand{\xmark}{\textcolor{purple}{$\boldsymbol{\times}$}}
\newcommand{\yes}{\textcolor{purple}{\checkmark}}

\usepackage[table]{xcolor}
\usepackage{booktabs}
\usepackage{colortbl}
\usepackage{multirow}
\usepackage{graphicx}
\usepackage{acronym}
\usepackage{arydshln} 
\usepackage{subcaption}
\usepackage{xspace}

\definecolor{lightblue}{HTML}{f3f7fc}
\definecolor{jPurple}{HTML}{6A2C91}
\definecolor{jMagenta}{HTML}{C2187A}
\definecolor{jLPink}{HTML}{F7C6DE}
\definecolor{jPinkBg}{HTML}{FDEEF5}
\definecolor{jLavBg}{HTML}{F1EAF9}

\newcommand{\rA}[1]{\cellcolor{jLPink}\textbf{#1}}
\newcommand{\rB}[1]{\cellcolor{jLPink!55!white}#1}
\newcommand{\rC}[1]{\cellcolor{jLPink!25!white}#1}
\newcommand{\rD}[1]{\underline{#1}}
\newcommand{\ours}[1]{\cellcolor{jPinkBg}#1}
\newcommand{\base}{\rowcolor{jLavBg}}

\newcommand{\ext}[2]{#1$^{#2}$}
\newcommand{\parsection}[1]{\noindent\textbf{#1:}}
\newcommand{\pcaimg}[1]{\includegraphics[width=\linewidth,trim={0 0 0 5cm},clip]{#1}}
\newcommand{\supmat}{supp.mat.\xspace}
\newcommand{\tabnote}[1]{$^{\text{#1}}$}
\makeatletter
\newcommand{\nolinkcite}[1]{{%
  \def\Hy@StartlinkName##1##2{}%
  \let\pdfendlink\relax
  \cite{#1}}}
\makeatother

\usepackage{listings}

\lstdefinestyle{pytorch}{
  language=Python,
  basicstyle=\small\ttfamily,
  breaklines=true,
  keepspaces=true,
  showstringspaces=false,
  frame=single,
  captionpos=t
}

\definecolor{cvprblue}{rgb}{0.21,0.49,0.74}
\usepackage[pagebackref,breaklinks,colorlinks,allcolors=cvprblue]{hyperref}

\title{J3PA: A Predictive Approach to Self-Supervised 3D Representation Learning on Point Clouds}

\author{
Pauline N\"{a}sslander$^{1,2*}$ \quad
Willem Verbeke$^{1*}$\quad
Adam Lilja$^{1,2}$ \quad
Fredrik Kahl$^{2}$ \quad
Lars Hammarstrand$^{2}$\\[0.5em]
$^{1}$Zenseact \qquad $^{2}$Chalmers University of Technology\\[0.3em]
{\small $^{*}$Equal contribution}\\[0.6em]
{\small\hypersetup{urlcolor=jPurple}
\href{https://huggingface.co/collections/Zenseact/j3pa}{Pretrained models} \quad$\cdot$\quad
\href{https://github.com/PaulineNasslander/J3PA}{Training code} \quad$\cdot$\quad
\href{https://research.zenseact.com/publications/j3pa/}{Project page}
}
}

\begin{document}
\twocolumn[{%
  \renewcommand\twocolumn[1][]{#1}%
  \maketitle
  \begin{center}
    \includegraphics[width=\textwidth]{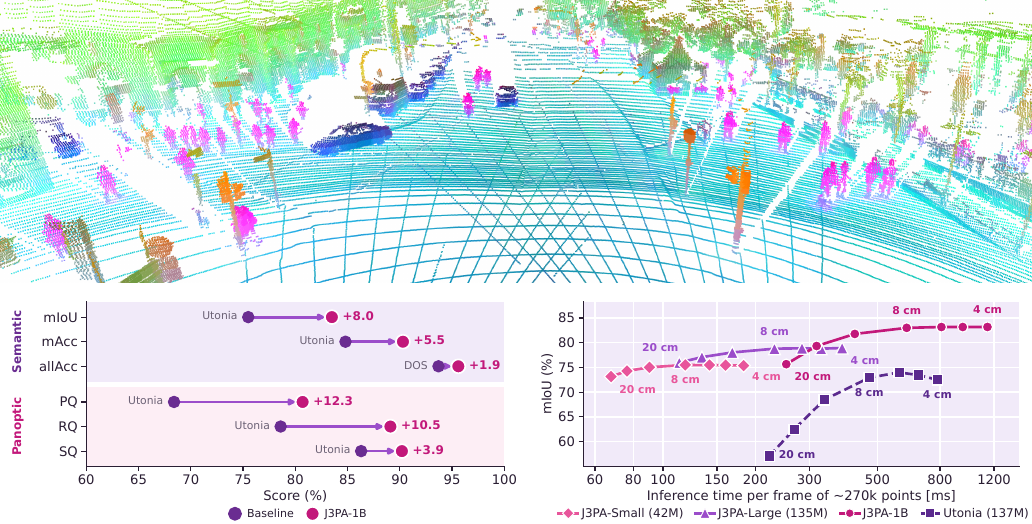}
    \captionof{figure}{J3PA learns robust features across a large range of inference budgets from only point clouds. Representations are visualized using PCA, where semantic concepts are clearly distinguishable and coherent. Linear probe results, reported on nuScenes, position J3PA as a competitive foundation model for outdoor point clouds.
    }
    \label{fig:title_fig}
  \end{center}
}]

\begin{abstract}
Learning versatile representations from unlabelled LiDAR point clouds is crucial for systems that perceive and interact with physical environments. While joint-embedding predictive architectures (JEPAs) have had great success in images and video, prior work on self-supervised learning (SSL) for point clouds has largely focused on invariance-based methods. We address positional information leakage when adapting JEPA for point clouds and introduce a sparse deformable attention layer to build a leakage-free predictor. To improve SSL at scale, we modify existing convolution-transformer hybrid encoder architectures into a pure transformer and train it on a combined corpus of outdoor point clouds. The resulting model, J3PA (Joint 3D-embedding Predictive Architecture), learns robust representations only from raw unlabeled point clouds without other modalities. Our largest model, J3PA-1B, is distilled into two variants, Large and Small with approximately 135M and 42M parameters respectively. Under linear probing, J3PA-1B markedly outperforms state of the art SSL methods and a fully supervised PTv3 across a large set of tasks and datasets, including a dataset unseen during training. We introduce a distillation procedure parametric in the voxel size to obtain models that are highly adaptable to different inference budgets and environments. Our models and training code are publicly released as LiDAR foundation models for autonomous driving. We believe our findings motivate predictive SSL as a viable path towards general foundation models for all point clouds.

\end{abstract}
    
\section{Introduction}
\label{sec:intro}

\acrodef{SSL}{self-supervised learning}
\acrodef{JEPA}{joint-embedding predictive architecture}
\acrodef{VFM}{Vision Foundation Model}
\acrodef{SOTA}{state of the art}
\acrodef{AD}{Autonomous Driving}
\acrodef{EMA}{Exponential Moving Average}
\acrodef{MHA}{Multi-Head Attention}
\acrodef{BEV}{Birds Eye View}
\acrodef{TTA}{Test Time Augmentations}
\acrodef{OOD}{out of distribution}
\acrodef{PCA}{Principal Component Analisys}

Models trained via \ac{SSL} have proven extraordinarily capable and flexible in a multitude of domains, across a wide array of tasks~\cite{dinov2, dinov3, gpt2, llama, deepseekllm}. These generic models can be used to solve previously challenging problems~\cite{dinov2}, enable downstream applications~\cite{depth_anything_v2}, and facilitate the development of new foundation models that solve more specific tasks~\cite{vggt_omega}. In addition, powerful generic representation spaces can form the basis for systems with zero-shot planning capabilities through world modeling~\cite{dino_wm, ad_jepa_planning}. While increasingly powerful foundation models have become readily available in image and video over the past years, point cloud encoders still lag behind. The recent work Utonia~\cite{utonia} made a significant step towards training a single generic encoder applicable to all point clouds, but it is still far smaller in parametric scale than contemporary image or video models~\cite{dinov3, vjepa_2p1}. Given the importance of LiDAR in embodied intelligent systems we see a need for stronger point cloud encoders. 

Top-performing \ac{SSL} models for LiDAR data are trained by learning invariances and/or distilling features from existing image-based models. Predictive video models have been shown to be competitive with the strongest invariance-based models~\cite{vjepa_2, vjepa_2p1}. Yet, predictive \ac{SSL} approaches based on the \ac{JEPA} paradigm remain under-explored for point clouds in comparison. In this paper, we show the feasibility of the predictive path towards foundation models for point clouds and introduce a model and training recipe, which we name \textbf{Joint 3D-embedding Predictive Architecture (J3PA)}. We do this by studying the predictor and supervision strategy, encoder architecture, and how to increase the flexibility of models under different inference budgets.

The limited body of work on predictive point cloud \ac{SSL} may partly stem from the difficulty in avoiding information leakage in the predictor. Unlike images, point clouds explicitly encode spatial information, making the masked prediction task trivial when conditioned on the masked point positions~\cite{asymmetric_self_distill_point}. We handle this through a leakage-free predictor that utilizes isolated queries in a new sparse deformable attention layer. Combined with our hierarchical masking scheme, this predictor allows the model to be supervised at multiple spatial scales, some far smaller than in earlier work~\cite{sonata}, without collapsing to simple geometric features.

A common trend in other fields is to develop purely transformer-based models, which tends to result in large performance gains when trained on sufficient data~\cite{dit, vggt, lvsm}. This has not been definitively shown for point clouds, where two-thirds of the parameters in \ac{SOTA} \ac{SSL} models come from submanifold convolutions~\cite{submanifold_convs}, while transformer blocks make up less than a third~\cite{ptv3}. We pursue the development of a pure transformer model at scale, focused on outdoor LiDAR data. 

Point cloud models are typically trained at a fixed voxel size. This presents a dilemma before any costly training, between small voxels, which give stronger dense-task performance at greater computational expense, and large voxels, which give cheaper but less capable models. Utonia~\cite{utonia} takes a step towards addressing this with domain-specific rescaling, but still uses a fixed effective voxel size per domain.
We alleviate the tradeoff by training models parametric in the the voxel size, enabling users to choose a voxel size at inference time and get performance equivalent to a model trained for just that configuration.

We find that J3PA outperforms current point cloud models by a large margin across \ac{AD} datasets, without any reliance on other modalities or vision foundation models. Our contributions are summarized as follows:
\begin{itemize}
    \item We show the potential of predictive \ac{SSL} and challenge current invariance-based techniques (\cref{fig:title_fig}). The predictive approach is made viable via a new leakage-free deformable attention predictor and a hierarchical masking scheme that allows supervision at multiple feature scales.
    \item We modify PTv3 into a pure transformer.
    \item We introduce a (self) distillation procedure parametrized by the voxel size, enabling single models to be used across a large array of inference budgets while retaining strong performance.
    \item We publicly release a family of performant LiDAR foundation models trained on a large corpus of outdoor scenes as well as corresponding training code.
\end{itemize}

\section{Related Work} \label{sec:rel_work}
Self-supervised learning for 3D point clouds draws on advances in image and video representation learning, but must account for the sparse and explicitly spatial nature of point cloud data.
We position our work within this literature, focusing on predictive learning and the geometric leakage that complicates its application to LiDAR point clouds.
We provide an extended discussion in \supmat~\ref{sup:extended_related_work}, including point transformers, deformable attention, and voxel-size generalization.

\parsection{SSL for images and video}
Invariance-based self-distillation methods learn representations by matching embeddings across augmented views~\cite{byol, dino}.
Predictive methods like \ac{JEPA} instead learn to infer representations of missing observations from available context~\cite{a_path_towards}.
This approach has been applied successfully to images~\cite{ijepa} and videos~\cite{vjepa, vjepa_2}, motivating our investigation of predictive \ac{SSL} for LiDAR point clouds.

\parsection{SSL for point clouds}
Point cloud representations can be learned by reconstructing masked inputs~\cite{georg_pointmae, pointmae, bosch_nomae}, predicting geometric targets such as occupancy~\cite{4docc, uno}, distilling features from vision foundation models~\cite{sal, lidar_clip, hilda}, or both~\cite{gasp}.
Other approaches learn directly in embedding space through contrastive learning~\cite{masked_scene_contrast} or self-distillation across augmented views~\cite{sonata, bosch_dos}.
Concerto~\cite{concerto} and Utonia~\cite{utonia} combine self-distillation with image foundation model distillation, while instance-aware methods improve the separation of individual objects~\cite{bosch_pointins, bosch_ghostpoint}.
Sonata~\cite{sonata} identifies geometric shortcuts as an obstacle to learning semantic representations and mitigates them by supervising coarser features.

\parsection{Predictive SSL for point clouds}
Prior work applies latent prediction to individual objects~\cite{point_jepa}, automotive scenes represented in dense \ac{BEV} grids~\cite{ad_l_jepa}, and future LiDAR observations~\cite{ad_jepa_wm}.
A challenge for prediction on sparse point clouds is that masked-token positions themselves reveal information about the missing geometry.
AsymDSD~\cite{asymmetric_self_distill_point} addresses this in object-level representation learning by preventing attention between masked queries.
Our work develops predictive \ac{SSL} for full outdoor point clouds using a sparse transformer encoder and a sparse deformable predictor whose queries independently attend to visible context. Combined with hierarchical masking, this enables supervision across spatial scales without exposing the configuration of masked queries to one another.

\section{Method}
\label{sec:method}
Our J3PA training methodology closely follows \acp{JEPA} on images~\cite{ijepa} and videos~\cite{vjepa, vjepa_2, vjepa_2p1}. We train two encoder models, teacher and student, simultaneously through an asymmetric distillation pipeline with a predictor. The teacher encodes complete point clouds, whereas the student processes cropped and masked views. One predictor per model stage estimates teacher embeddings at masked locations given the encoded student view and the location of the missing features. Student and predictor weights are updated via stochastic gradient descent. During self-distillation, the teacher is updated as an \ac{EMA} of the student encoder. Our largest model, J3PA-1B, follows this procedure while two smaller models J3PA-Large and J3PA-Small are trained by distilling the final J3PA-1B model as a fixed teacher. Detailed training hyperparameters can be found in \supmat~\ref{supmat:training-details}.

\begin{figure*}[h]
    \centering
    \includegraphics[width=\linewidth]{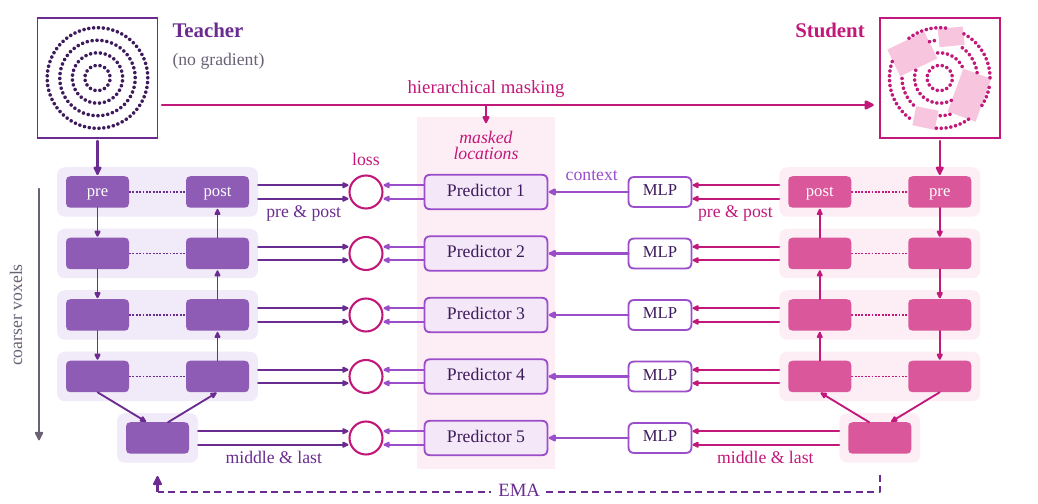}
    \caption{J3PA training pipeline. A teacher encoder processes the full point cloud while a student encoder processes a partial observation of the scene. Pre- and post-U-Net student features are combined into stage-wise context features. A separate predictor per stage processes the context to predict teacher embeddings in masked segments. The loss is the $\operatorname{L1}$ distance between true and estimated teacher features. The student and predictors are updated via gradient descent, whereas the teacher receives no gradient and is an \ac{EMA} of the student.}
    \label{fig:overview}
\end{figure*}

\Cref{fig:overview} outlines the J3PA methodology. \Cref{sec:encoder} describes the encoder, \cref{sec:masked_prediction} the masking strategy and training objective, \cref{sec:sparse_deformable_predictor} the predictor, and \cref{sec:voxelsize_parameterization} how the model is generalized to different voxel sizes.

\subsection{Encoder}
\label{sec:encoder}
Our encoder is a sparse point transformer with multiple resolution stages and a U-Net~\cite{unet} structure (referred to as U-Net below), based on PTv3~\cite{ptv3}. We make improvements to the usage of RoPE and remove sparse convolutions.
Prior \ac{SSL} work discards the U-Net after identifying that it results in models focusing on low-level geometric features~\cite{sonata, concerto, utonia}. 
In contrast, our training strategy allows us to retain the U-Net without any indication of semantic information degradation and an increase in downstream performance at the same parameter count (\cref{sec:ablate_supervision_stages}).

\parsection{Pure transformer-based model}
Most contemporary \ac{SSL} work uses the PTv3-encoder~\cite{sonata, bosch_dos, bosch_ghostpoint, bosch_pointins, concerto, utonia} which employs sparse convolutions that enable the model to resolve fine-grained relative positions. These convolutions make up over two-thirds of the parameters in PTv3-based models. Inspired by the success of pure transformer architectures in other domains when trained at scale~\cite{vit}, we pursue a \emph{pure} transformer-based architecture for point clouds that outperforms PTv3 for \ac{SSL} at scale.

\parsection{Rotary Position Embedding design} 
To retain fine-grained spatial awareness in a transformer, we take inspiration from~\cite{litept, utonia} and extend RoPE~\cite{rope} to 3D positions. We divide the embedding dimensionality into three sub-parts and apply standard 1D RoPE on the coordinates in each subspace. Rather than using metric coordinates we compute RoPE's input coordinates in voxel units so it can resolve individual voxel positions. Second, we use the full channel width of our model to build the RoPE frequency basis rather than using the same basis for each attention head. Angles for different frequencies and coordinate axes are interleaved so that each attention head has low- and high-frequency positional encodings and is sensitive to all three axes (see \supmat~\ref{supmat:supp_rope}).  Our RoPE implementation can be used with conventional, GPU-optimized, channel counts rather than multiples of 6 without loss of performance, speeding up inference. 
Combined, this allows the fine model stages to be sensitive to position without sparse convolutions yielding faster, stronger models. 
We make further implementation and model refinements to our training pipeline compared to PTv3 and the Pointcept repository~\cite{pointcept} detailed in \supmat~\ref{supmat:subtle_model_differences} and the released code.

\subsection{Masked Prediction}
\label{sec:masked_prediction}

\parsection{Hierarchical masking strategy}
Intuitively, coarser model stages with a larger receptive field should capture semantic concepts that cover larger spatial scales while earlier stages focus on fine-grained geometry and small objects. This is reinforced directly with our hierarchical masked prediction objective.
\begin{figure}
    \centering
    \includegraphics[width=0.75\linewidth]{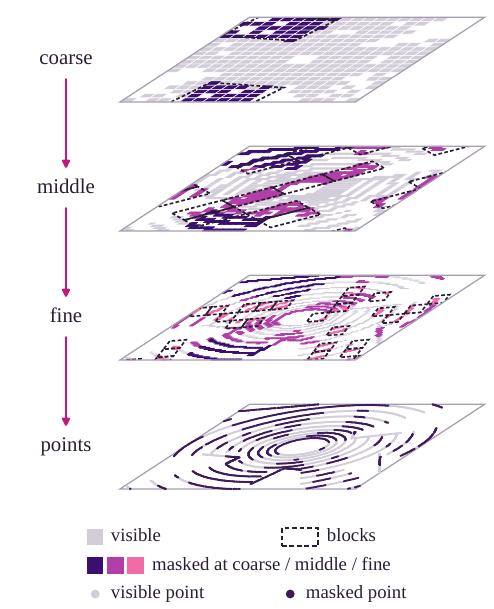}
    \caption{The hierarchical masking strategy illustrated in \ac{BEV} for a three-stage model. At each level, we sample new (primary) 3D blocks proportional to the stage voxel size. The sampling starts at the coarsest level and maps the existing masks to voxels at finer resolutions to eliminate information leakage in masked content. New primary masks at each stage are chosen among the set of visible voxels after the application of coarser masks.}
    \label{fig:masking_strategy}
\end{figure}
\Cref{fig:masking_strategy} exemplifies our masking strategy. For each scene and resolution stage, we randomly choose a 3D cuboid mask size $c$. The space is then divided into a randomly-oriented grid of $c$-sized blocks and we sample blocks to be masked until the points in the masked blocks exceed a predetermined fraction of the remaining points in that scene. A consistent fraction of information is retained after masking for all scenes. Coarser stages with larger voxel sizes have larger expected mask sizes than fine-voxel stages. Masks are sampled at the coarsest resolution first and cascaded to the input point cloud by iteratively masking all finer stage voxels inside a masked coarser voxel. This ensures that there is no information leakage. The mask cascade is similar the approach in~\cite{bosch_nomae}, while the information-weighted rotated cuboid sampling is, to the best of our knowledge, novel.

We follow~\cite{ijepa} and sample a context window along each spatial axis with a random origin. The student only sees unmasked points inside the context window, and prediction is limited to masked tokens in the window. This does not increase peak model performance, but reduces the risk of collapse.

\parsection{Primary masks} The 3D cuboid mask-blocks sampled for a particular stage are called primary masks of the stage. Masked voxels are categorized by whether they lie inside primary masks or originate from a mask propagated from a coarser model stage. Queries for supervising the model are only sampled for occupied voxels inside primary masks. Using the primary masks adapts the task to the receptive field of each stage of the model.

\parsection{Supervision signal}
Teacher features both before and after the U-Net are predicted, given the student outputs before and after the U-Net. For the deepest model stage we use the middle and last layer outputs of that stage instead. This is inspired by~\cite{vjepa_2p1} where supervising intermediate ViT stages gives improved results on dense tasks. Both student outputs for each stage are combined into a single context with an MLP. The predictor uses that context to predict both teacher embeddings at each stage.

For prediction and supervision, we sample a subset, $\mathcal{Q}_s$, of masked voxel tokens as queries, $q$, for each stage, $s$. We let the cardinality $|\mathcal{Q}_s|$ be proportional to the number of occupied voxels in a given sample and stage, but enforce $|\mathcal{Q}_s|\geq 1$ per sample. 
Our loss function is defined as
\begin{equation}
    \mathcal{L}_\text{J3PA} = \sum_{s=1}^{S} \sqrt{\tfrac{D_s}{D_{5}}} \left( \ell_s + \ell'_s \right)\\
\end{equation}
\begin{equation}
    \ell_s = \frac{1}{|\mathcal{Q}_s|\cdot D_s}\sum_{q \in \mathcal{Q}_s}\sum_{d}^{D_s}
    \big\| \hat{f}_{s,q,d} - \operatorname{LN}(\operatorname{sg}(f_{s,q,d})) \big\|_1
\end{equation}
with $D_s$ the channel count at stage $s$, $\ell_s$ the loss in the prediction ($\hat{f}_{s,q,d}$) of the post-U-Net target ($f_{s,q,d}$). Analogously, $\ell'_s$ is the loss for the pre-U-Net target. The stage-wise scaling $\sqrt{D_s/D_{5}}$ equalizes the expected gradient norm between stages (see \supmat~\ref{supmat:loss_scaling}). Losses are computed separately for each rank to avoid inter-GPU communication during the forward pass.

\parsection{Independent teacher and student augmentations}
Absolute orientation and position dependence of features is undesirable for downstream tasks. We remedy this by sampling separate rotation and translation augmentations for student and teacher inputs. The masked positions used to query the predictor are given in the student coordinate system, but retain the teacher embeddings as the target. Absolute position and orientation become unpredictable from the context, forcing the model to encode relative geometry and semantics. 
The teacher input orientation can be revealed to the student more subtly through the sampled masks and context window if applied axis-aligned to the teacher voxel grid. For this reason, mask blocks are randomly rotated for each sample and model stage as shown in~\cref{fig:masking_strategy}.

\subsection{Sparse Deformable Predictor}
\label{sec:sparse_deformable_predictor}

\begin{figure*}[t]
    \centering
    \includegraphics[width=\linewidth]{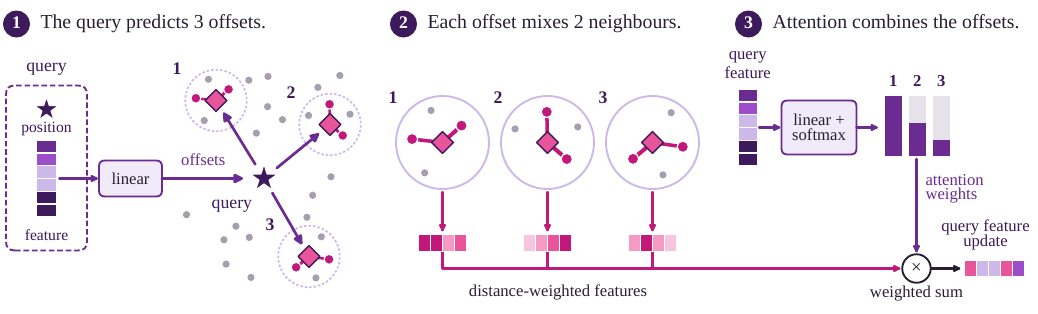}
    \caption{Our sparse deformable attention layer learns where to look for relevant information. The layer first linearly estimates $O=3$ offsets from the query feature. For each offset, the features of its k=2 nearest neighbors are aggregated through a weighted sum. The weights are a softmax over a linear projection of each neighbor's direction and distance to the sampled location. Crucially, this makes the aggregation differentiable with respect to the offsets. An attention-weighted sum of the $O$ distance features, with attention scores predicted from the query, forms the output.}
    \label{fig:sparse_def_pred}
\end{figure*}

\acp{JEPA} for vision use ViTs~\cite{vit} as the predictor, where mask tokens are processed together with the student features in a series of self-attention blocks~\cite{ijepa,vjepa}. For point clouds, the existence of a point at a particular location \textit{is} the information. A predictor based on self-attention that processes several mask tokens simultaneously therefore leaks information about the masked data regardless of the student context, weakening the incentive to improve context embeddings~\cite{pointmae, asymmetric_self_distill_point}.
The natural solution is a series of cross-attention blocks processing each query individually. Naive implementation leads to sub-optimal context groups or high memory and compute requirements. We introduce a stronger and more efficient predictor design that collects information in an adaptive spatial neighborhood around query locations.

\parsection{Sparse deformable attention} 
We build on~\cite{deformable_detr} and introduce a sparse deformable attention layer (\cref{fig:sparse_def_pred}). Each query consists of a position and a feature vector. Deformable attention uses the query feature to predict $O$ offsets per attention head from the query location. We then find the $k$-nearest student context features for each offset and combine them into a single feature vector with a distance-dependent learnable mixing. This step ensures the existence of gradients in the offset positions. The update to the query feature is an attention-weighted sum of the offset features with attention scores predicted from the query feature. In our training, we use $O=3$ and $k=2$ and find that increasing these values is not necessary for strong model performance.

Offset prediction biases are initialized spherically across heads with a radius dependent on the voxel size of the model stage the predictor acts on. Deformable attention layers are interleaved with feedforward layers into pre-normalized deformable attention blocks. We find that our sparse deformable cross-attention blocks boost the downstream performance and convergence speed of our \ac{JEPA} by a large margin, while being cheap in both memory and compute (\cref{sec:ablate_predictor}).

\parsection{Mask-size-conditioned predictor}
The training of J3PA-1B uses a fixed voxel size and mask blocks with sizes independently drawn per stage for each data sample. The granularity of embedded predictions should depend on the mask size. We therefore use FiLM layers~\cite{film} to condition the predictor on mask size embeddings in self-distillation trainings with inputs from a learned embedding table. This stabilizes self-distillation trainings but does not affect peak downstream performance.

\subsection{Generalizing to Different Voxel Sizes}
\label{sec:voxelsize_parameterization}

\parsection{Precision vs efficiency} The voxel size used in a point cloud encoder provides a clear tradeoff. Small voxel sizes result in large compute and memory footprints while providing the resolution necessary to identify small objects. Models trained at a fixed voxel size tend to generalize poorly to other voxel sizes at inference. We show this for the strongest \ac{SSL} baseline evaluated in~\cref{sec:experiments}. To facilitate broader application, we introduce a training strategy that results in a single trained model generalizing to a range of voxel sizes at inference.

\parsection{Conditioning encoder and predictor on the voxel size} During distillation of our smaller models we draw random voxel sizes for each data sample. The teacher model always uses a fixed voxel size to generate the prediction targets. We input the voxel size as a feature for all points and use a stage-wise AdaLN-single~\cite{adaln_single} conditioning of the attention blocks on a sample-wise shared voxel size embedding. In the predictor, we concatenate and process the voxel-size embedding and mask-size embedding through an MLP whose output is used for AdaLN~\cite{dit} conditioning. Voxel size embeddings are a Fourier expansion of the normalized log voxel size. The procedure results in model outputs that are a function of the input point cloud and the specified voxel size, trained to predict fixed-difficulty targets.

In~\cref{sec:ablate_voxel_size_param} we show that our voxel-size parametrization also works during self-distillation. However, when scaling to the size of J3PA-1B, we found it to cause instabilities in its current form. We presume that this procedure could be useful for handling density differences when training on combined sets of indoor and outdoor point clouds without inconsistently representing the true scale of objects, but we leave that for future work.

\begin{table*}[h]
\centering
\caption{
Semantic and panoptic segmentation performance on the validation set for each dataset. Our \ac{SSL} trainings exclude the Waymo Open Dataset ($\dagger$ out-of-distribution) and validation scans for any dataset. $P$, and $U$ denotes results reported by PointINS~\cite{bosch_pointins} and Utonia~\cite{utonia} respectively. $^*$ Evaluation of released model. Best baseline is underlined and ranking indicated \colorbox{jLPink}{First}, \colorbox{jLPink!55!white}{Second}, and \colorbox{jLPink!25!white}{Third}.}
\label{tab:outdoor-segmentation}
\begingroup
\footnotesize
\setlength{\tabcolsep}{3pt}
\renewcommand{\arraystretch}{1.1}
\resizebox{\textwidth}{!}{%
\begin{tabular}{llc*{14}{c}}
\toprule
& & \multicolumn{6}{c}{nuScenes~\nolinkcite{nuscenes}}
& \multicolumn{6}{c}{SemanticKITTI~\nolinkcite{semantic_kitti, semantic_kitti_base}}
& \multicolumn{3}{c}{Waymo~\nolinkcite{waymo}} \\
\cmidrule(lr){3-8}\cmidrule(lr){9-14}\cmidrule(lr){15-17}
& & \multicolumn{3}{c}{Sem. seg.} & \multicolumn{3}{c}{Pan. seg.}
& \multicolumn{3}{c}{Sem. seg.} & \multicolumn{3}{c}{Pan. seg.}
& \multicolumn{3}{c}{Sem. seg.} \\
\cmidrule(lr){3-5}\cmidrule(lr){6-8}\cmidrule(lr){9-11}\cmidrule(lr){12-14}\cmidrule(lr){15-17}
Methods & Params & mIoU & mAcc & allAcc & PQ & RQ & SQ & mIoU & mAcc & allAcc & PQ & RQ & SQ & mIoU & mAcc & allAcc \\
\midrule
\color{gray}PTv3 (fully sup.) & \color{gray}46M & \color{gray}80.4 & \color{gray}87.2 & \color{gray}94.7 & \color{gray}\ext{69.9}{P} & \color{gray}\ext{80.5}{P} & \color{gray}\ext{86.3}{P} & \color{gray}70.8 & \color{gray}76.1 & \color{gray}92.6 & \color{gray}\ext{58.2}{P} & \color{gray}\ext{67.7}{P} & \color{gray}\ext{78.7}{P}& \color{gray}71.3 & \color{gray}80.5 & \color{gray}94.7 \\
Sonata (lin.) & 108M & 66.1 & 77.2 & 92.4 & \ext{50.7}{P} & \ext{61.6}{P} & \ext{79.8}{P} & 62.0 & 72.5 & 91.0 & \ext{34.5}{P} & \ext{44.2}{P} & \ext{65.5}{P} & 60.5 & 72.5 & \rD{92.5} \\
DOS (lin.) & 108M & 74.1 & \rD{84.8} & \rD{93.7} & \ext{57.4}{P} & \ext{68.5}{P} & \ext{82.8}{P} & 67.5 & \rD{78.1} & -- & \ext{49.6}{P} & \ext{60.9}{P} & \ext{71.8}{P} & \rD{66.1} & \rC{\rD{77.1}} & -- \\
PointINS (lin.) & 108M & 74.4 & -- & -- & 62.2 & 72.8 & 84.5 & 66.9 & -- & -- & 52.8 & 63.2 & 73.7 & -- & -- & -- \\
GhostPoint (lin.) & 108M& 74.2 & -- & -- & 62.8 & 73.2 & 84.6 & -- & -- & -- & -- & -- & -- & -- & -- & -- \\
Concerto (lin.) & 137M & \ext{74.2}{U} & \ext{83.2}{U} & \ext{93.3}{U} & -- & -- & -- & \ext{66.6}{U} & \ext{76.2}{U} & \ext{91.4}{U} & -- & -- & -- & \ext{62.2}{U} & \ext{74.4}{U} & \ext{89.4}{U} \\
Utonia (lin.) & 137M & \rD{75.5} & \rD{84.8} & \rD{93.7} & \rD{68.4}\tabnote{*}  & \rD{78.6}\tabnote{*}  & \rD{86.3}\tabnote{*}  & \rD{67.7} & 76.9 & \rD{91.9} & \rC{\rD{59.7}}\tabnote{*} & \rD{69.1}\tabnote{*} & \rD{76.3}\tabnote{*} & 63.8 & 75.4 & 89.9 \\
\ours{J3PA-Small (lin.)} & \ours{42M} & \rC{76.3} & \rC{85.9} & \rC{94.9} & \rC{74.2} & \rC{83.2} & \rC{88.6} & \rC{70.0} &	\rC{78.5} &	\rB{93.3} &	59.5 &	\rC{69.2} &	\rA{79.5}  & \rC{66.9}$^{\dagger}$ & 76.7$^{\dagger}$	& \rC{94.2}$^{\dagger}$  \\
\ours{J3PA-Large (lin.)} & \ours{135M} & \rB{79.9} & \rB{88.1} & \rB{95.3} & \rB{77.3} & \rB{86.3} & \rB{89.1} & \rB{71.3} & \rB{78.6} & \rC{93.2} & \rB{61.5} & \rB{71.0} & \rC{76.9} & \rB{67.9}$^{\dagger}$ & \rB{77.5}$^{\dagger}$ & \rB{94.4}$^{\dagger}$
  \\
\ours{J3PA-1B (lin.)} & \ours{1.0B} & \rA{83.5} & \rA{90.3} & \rA{95.6} & \rA{80.7} & \rA{89.1} & \rA{90.2} & \rA{73.1} & \rA{79.1} & \rA{93.5} & \rA{64.3} & \rA{73.6} & \rB{77.5} & \rA{72.0}$^{\dagger}$ & \rA{81.3}$^{\dagger}$ & \rA{94.9}$^{\dagger}$
\\
\bottomrule
\end{tabular}%
}
\endgroup
\end{table*}
\section{Experiments}
\label{sec:experiments}
In this section, we quantitatively show the strength of J3PA and motivate design choices.

\parsection{Implementation details} Our largest model J3PA-1B is trained for $\sim100$k steps, or 11 days, on 32 A100 GPUs. Distillation of our smaller models follows the same overall procedure with a fixed J3PA-1B teacher. We add the voxel size parameterization procedure to the distillation training. See \supmat~\ref{supmat:training-details} for a detailed hyperparameter overview.

\parsection{Pre-training data}
Our family of models are trained on a corpus of autonomous driving datasets specified in \supmat~\cref{tab:data-splits}. 
One of the goals of \ac{SSL} is to learn robust features that generalize to new datasets and applications. Therefore, the test and validations sets are not included for any labeled dataset, unlike common practice in point-cloud \ac{SSL}~\cite{sonata, concerto, utonia}.
In addition, we do not use Waymo Open Dataset~\cite{waymo} in the \ac{SSL} training of our models and show on its validation set that our model can achieve \ac{SOTA} performance despite being out of distribution. 

\subsection{Evaluating representation quality}
\label{sec:results}

We evaluate our learned representations by training a linear probe on a frozen encoder for semantic and panoptic segmentation on several driving datasets. Probes are trained separately on each dataset. 
Following earlier \ac{SSL} works, we mimic the Pointcept evaluation protocol~\cite{pointcept} for semantic segmentation (See \supmat~\ref{supmat:eval}). For panoptic segmentation we adopt the PointGroup procedure~\cite{pointgroup} as in~\cite{bosch_ghostpoint, bosch_pointins, bosch_dos}. The probe is linear in the semantic classification head, while the offset prediction used for instance grouping is a 2-layer MLP.  

The main results are reported in \cref{tab:outdoor-segmentation}. Our largest model J3PA-1B achieves \ac{SOTA} results on all probed datasets and tasks by significant margins. In panoptic segmentation, the gap with previous methods is especially large, and we speculate that the predictive nature of our models has an advantage compared to invariance methods for distinguishing object instances. The Large and Small variants are also competitive with the best baselines on all benchmarks. We stress that our models did not see any Waymo Open Dataset scans during \ac{SSL} training, yet outperform previous methods with a frozen encoder. More datasets are benchmarked in \supmat~\ref{supmat:other_dataset}.

\begin{table}[t]
\centering
\caption{Semantic and panoptic segmentation on nuScenes~\cite{nuscenes} for a self-distilled J3PA with 131M parameters and trained only on nuScenes (train set) scans. $^*$ Evaluation of released model.}
\label{tab:less_param_comp}
\begingroup
\footnotesize
\setlength{\tabcolsep}{3pt}
\renewcommand{\arraystretch}{1.1}
\begin{tabular}{lc*{6}{c}}
\toprule
& & \multicolumn{3}{c}{Sem. seg.} & \multicolumn{3}{c}{Pan. seg.} \\
\cmidrule(lr){3-5}\cmidrule(lr){6-8}
Methods & Params & mIoU & mAcc & allAcc & PQ & RQ & SQ \\
\midrule
Utonia (lin.) & 137M & 75.5 & 84.8 & 93.7 & 68.4\tabnote{*}  & 78.6\tabnote{*}  & 86.3\tabnote{*} \\
\ours{J3PA (lin.)} & \ours{131M} & \rA{78.3} & \rA{85.8} & \rA{95.0} & \rA{74.1} & \rA{83.6} & \rA{88.0} \\
\bottomrule
\end{tabular}
\endgroup
\end{table}

\Cref{tab:less_param_comp} shows the strength of our proposed training recipe when self-distilling using only nuScenes. Our tested model has a comparable parameter count to the strongest evaluated baseline. When trained at a smaller scale on a single dataset, J3PA gives competitive results. 

\begin{figure}
    \centering
    \includegraphics[width=\linewidth]{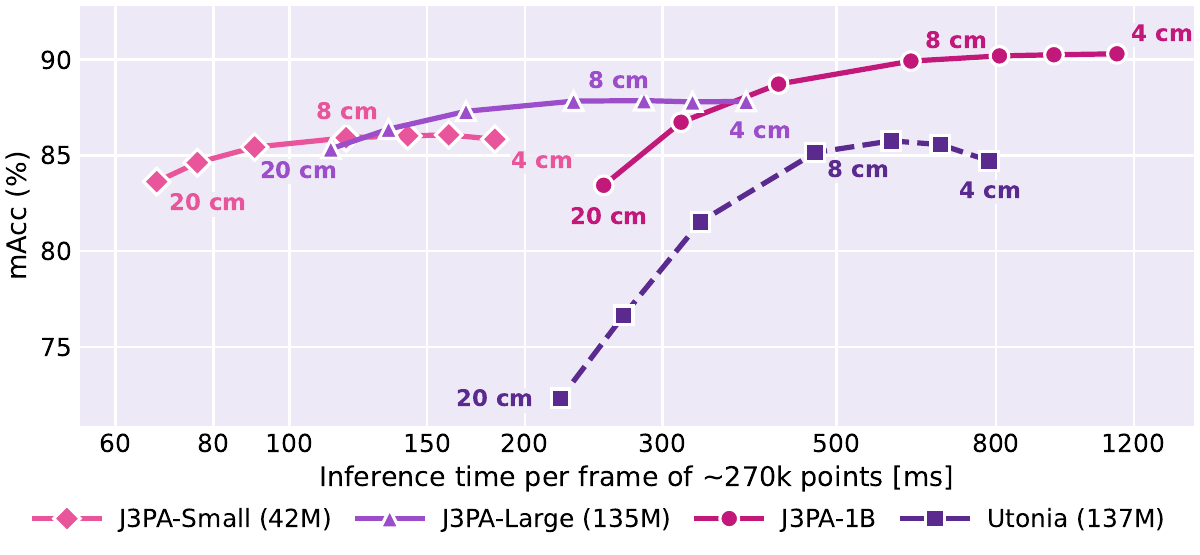}
    \caption{Linear probe semantic segmentation mean accuracy performance on nuScenes over inference time for our family of models vs. Utonia evaluated in our pipeline.}
    \label{fig:macc_over_inf_time}
\end{figure}

\parsection{Accuracy vs. Inference Time}
The flexibility of our family of models for different applications is strengthened by the voxel size parameterization and generally high inference throughput. \cref{fig:title_fig} and \ref{fig:macc_over_inf_time} show performance as a function of inference time on a single A100 GPU. This demonstrates that the distilled models are robust to different voxel sizes, and one can choose configurations that are very efficient without sacrificing performance compared to earlier work.

\parsection{Qualitative}
A qualitative assessment of the encoded representations from J3PA-1B can be made from the \ac{PCA} visualization in~\cref{fig:title_fig}. Further \ac{PCA} visualization for each stage of the model can be found in~\supmat~\ref{supmat:qualitative}.

\subsection{Ablations}
\label{sec:ablations}
We use the Zenseact Open Dataset (ZOD)~\cite{zod} in all ablations due to its large scene diversity. In addition, developing the method on ZOD prevents accidental over-optimization on common benchmark datasets reported in \cref{sec:results}. Since ZOD only contains 3D bounding boxes in the 120 degree field of view of the forward-looking camera, we obtain point-wise labels by assigning each point the class of the object it falls inside, or background otherwise. We impose no range limit on the point clouds. Note that this design results in a \textit{things}-only benchmark without large \textit{stuff} classes. 

For computational tractability, all experiments follow a 100-epoch self-distillation training scheme on a $78$M parameter J3PA encoder model with a finest stage voxel size with $10$ $\text{cm}$ sides. We start the teacher EMA at $0.992$ and linearly increase it to $0.999$ over $5$k training steps. All other training hyperparameters are equivalent to our main trainings. To report probe results on ablation models, we use a simplified evaluation protocol and train all probes for $5000$ steps. We evaluate at regular intervals during \ac{SSL} training and report peak numbers for all experiments.

\parsection{Supervision stages}
\label{sec:ablate_supervision_stages}
Our base model keeps the U-Net structure originally introduced in PTv3 and predicts teacher features before and after the U-Net. For the deepest model stage we predict the middle and last layer representations. These choices are ablated in \cref{tab:abl-supervision}. Predicting only stages $3$ to $5$ is most similar to earlier works~\cite{sonata, utonia}, but predicting all model stages gives better results. Having a U-Net gives better results than a model without it at the same parameter count, with both configurations supervising only the last layer of each stage. Supervising both pre- and post-U-Net features gives the best results.

\begin{table}[t]
\centering
\caption{Sem. seg. for different supervision targets.}
\label{tab:abl-supervision}
\footnotesize
\begin{tabular}{@{}lccc@{}}
\toprule
Variant & mAcc & mIoU & allAcc \\
\midrule
\base Base (All stages up and down path) & \textbf{50.7} & \textbf{41.2} & \textbf{84.6} \\
Up path only, all stages & 49.3 & 39.2 & 83.5 \\
Up path only, S5/S4/S3 & 47.9 & 37.7 & 82.0 \\
Down path only, all stages & 47.8 & 38.3 & 82.4 \\
Down path only, S5 & \multicolumn{3}{c}{\textit{collapsed}} \\
\bottomrule
\end{tabular}
\end{table}

\parsection{Predictor}
\label{sec:ablate_predictor}
We evaluate three predictor designs in \cref{tab:abl-predictor}. In the self-attention predictor, all mask and context tokens interact, causing information leakage. The Cross-attention predictor is leakage-free since each query attends to a group of context tokens individually. The context group is determined per query based on space-filling curves and we iterate through multiple curves to improve the neighborhood coverage. The results clearly indicate that a leakage-free predictor is necessary for learning strong representations. Our sparse deformable attention predictor boosts performance significantly while being faster and less memory-intensive to train.

\begin{table}[t]
\centering
\caption{Sem. seg. for different predictor variants. Measured with batch size 4 and approximately 270k points per frame. Parameter counts are predictor-only.}
\label{tab:abl-predictor}
\footnotesize
\setlength{\tabcolsep}{3pt}
\resizebox{0.95\linewidth}{!}{
\begin{tabular}{@{}lcccc@{}}
\toprule
Variant & \shortstack{Pred. \\ Params (M)} & mAcc/mIoU/allAcc
        & \shortstack{fwd.+bwd. latency\\(ms/layer) }
        & \shortstack{Peak mem.\\(GB) } \\
\midrule
\base \shortstack{ Base \\ (Sp. def. attn.)} & 98.09 & \textbf{50.7} / \textbf{41.2}/ \textbf{84.6} & 90.0  & 1.27 \\
Cross-attention & 104.89 & 43.1 / 34.0 / 79.7 & 206.2 & 6.81 \\
Self-attention  & 104.88 & 29.2 / 22.2 / 66.0 & 189.7 & 7.12 \\
\bottomrule
\end{tabular}
}
\end{table}

\parsection{RoPE and sparse convolutions}
\label{sec:ablate_rope}
We evaluate the effect of different RoPE configurations and the inclusion of sparse convolutions in \cref{tab:abl-rope}.
We see that scaling RoPE input positions to be in voxel units for each stage notably improves performance. Using downscaled metric coordinates as in Utonia yields poor performance when used without convolutions as fine model stages struggle to distinguish nearby voxels. Using the same rotation matrices in each attention head works well. Applying different rotations along the full channel width requires careful interleaving of the RoPE matrix so each attention head gets low- and high-frequency positional encodings but results in increased final performance (More details in \supmat~\ref{supmat:supp_rope}).

In PTv3-based models each attention block is preceded by convolutions, which make up more than two-thirds of the parameters. Convolutions are sensitive to single-voxel position differences by construction. \cref{tab:abl-rope} shows that models with convolutions are more robust w.r.t. the RoPE configuration than pure transformers. However, they lag behind transformers ability to distinguish positions with sufficient granularity. We expect the advantage of the transformer to grow with parameters and compute as in our J3PA-1B training, but have not trained any model with convolutions at that scale.

\begin{table}[t]
\centering
\caption{Sem. seg. for different RoPE and convolution configurations. All models have equivalent parameter counts.}
\label{tab:abl-rope}
\footnotesize
\begin{tabular}{@{}lcccc@{}}
\toprule
RoPE & Conv. & mAcc & mIoU & allAcc \\
\midrule
\base Base (pos.\ scale, interleave) & \xmark & \textbf{50.7} & \textbf{41.2} & \textbf{84.6} \\
Head-wise, with pos.\ scale & \xmark & 50.1 & 40.4 & 84.0 \\
Full-channels no pos.\ scale & \xmark & 46.7 & 37.8 & 82.6 \\
Full-channels no interleave & \xmark & 40.5 & 31.5 & 76.8 \\
Utonia RoPE & \xmark & 34.8 & 26.3 & 70.5 \\ \hline
Base & \yes & 47.0 & 38.5 & 82.9 \\
Utonia RoPE & \yes & 47.3 & 38.5 & 83.0 \\
No RoPE & \yes & 46.0 & 37.7 & 82.3 \\
\bottomrule
\end{tabular}
\end{table}

\parsection{Self-distillation with voxel size conditioning}
\label{sec:ablate_voxel_size_param}
Including the voxel size parameterization in self-distillation is possible and leads to a model more robust to a range of voxel sizes compared to default ablation model, see \cref{tab:abl-voxel-param}.

\begin{table}[t]
\centering
\caption{Sem. seg. voxel size parameterization on ZOD.}
\label{tab:abl-voxel-param}
\footnotesize
\begin{tabular}{@{}lcc@{}}
\toprule
Trained voxel size & \shortstack{mAcc/mIoU/allAcc\\@ 4cm}
        & \shortstack{mAcc/mIoU/allAcc\\@ 20cm}\\
\midrule
\base Base, VS $=10$cm & 65.0/\textbf{46.3}/\textbf{87.2} & 57.0/39.3/83.3 \\
VS $\in[4, 20]$cm   & \textbf{65.6}/46.1/\textbf{87.2} & \textbf{59.2}/\textbf{41.4}/\textbf{85.1} \\
VS $=20$cm          & --/--/-- & 46.9/30.7/77.1 \\
\bottomrule
\end{tabular}
\end{table}

\section{Conclusion}
\label{sec:conclusion}
This paper presents J3PA, a self-supervised training recipe for point clouds. A leakage-free predictor built with sparse deformable attention layers and a pure transformer-based encoder model unlock the full potential of \ac{JEPA} for point clouds. We train a family of foundation models at scale on a corpus of \ac{AD} LiDAR data.

Our models achieve \ac{SOTA} results across semantic and panoptic segmentation benchmarks on all evaluated outdoor datasets. This includes the Waymo Open Dataset~\cite{waymo} which was not used during the training of our \ac{SSL} models. Meanwhile, an improved distillation recipe makes our models highly robust to the choice of voxel size at inference. This results in adaptive and favorable performance-efficiency tradeoffs compared to previous models.

\parsection{Future and limitations}
Our method has only been developed and tested on \ac{AD} scenes as such datasets generally contain large quantities of dense LiDAR clouds. However, we see no inherent limitations of the method to this domain. Extending the method to a broader set of LiDAR data from other robot embodiments, or training jointly on indoor and outdoor scenes would be a natural next step. The presented voxel-size parameterization might have the potential to enable joint trainings without manually tuning scaling factors that might confuse spatial scales of objects. We leave this for future work.

\acrodef{LLM}{Large Language Model}

\section{Acknowledgements}

We thank Luca Caltagirone and Anni Carl\'{e}n for insightful discussions throughout this project. Jonas Andersson and Martin Nilsson made the distributed training of our largest models possible, and we appreciate their tireless support with GPU infrastructure and dataset storage. 
We are also grateful to Erik Werner for discussions on potential symmetries of the JEPA loss function, and for thorough feedback on the manuscript.  Maciej Wozniak provided further valuable feedback on the text.

This work was partially supported by the Wallenberg AI, Autonomous Systems and Software Program (WASP) funded by the Knut and Alice Wallenberg Foundation. Additional computational resources were provided by the National Academic Infrastructure for Supercomputing in Sweden (NAISS) on Arrhenius, funded by the Swedish Research Council, and on the Berzelius system, funded by the Knut and Alice Wallenberg Foundation and operated by NAISS.

{
    \small
    \bibliographystyle{ieeenat_fullname}
    \bibliography{main}

@String(CVPR= {IEEE Conf. Comput. Vis. Pattern Recog.})

@String(ICCV= {Int. Conf. Comput. Vis.})

@String(ECCV= {Eur. Conf. Comput. Vis.})

@String(ICLR = {Int. Conf. Learn. Represent.})

@String(AAAI = {AAAI})

@String(CVPR  = {CVPR})

@String(ICCV  = {ICCV})

@String(ECCV  = {ECCV})

@String(ICLR  = {ICLR})

@inproceedings{utonia,
    title={Utonia: Toward One Encoder for All Point Clouds},
    author={Yujia Zhang and Xiaoyang Wu and Yunhan Yang and Xianzhe Fan and Han Li and Yuechen Zhang and Zehao Huang and Naiyan Wang and Hengshuang Zhao},
    booktitle={Forty-third International Conference on Machine Learning},
    year={2026},
    url={https://openreview.net/forum?id=f2CGTzHQ4E}
}

@inproceedings{sonata,
    title={Sonata: Self-Supervised Learning of Reliable Point Representations},
    author={Wu, Xiaoyang and DeTone, Daniel and Frost, Duncan and Shen, Tianwei and Xie, Chris and Yang, Nan and Engel, Jakob and Newcombe, Richard and Zhao, Hengshuang and Straub, Julian},
    booktitle={CVPR},
    year={2025}
}

@inproceedings{ptv3,
    title={Point Transformer V3: Simpler, Faster, Stronger},
    author={Wu, Xiaoyang and Jiang, Li and Wang, Peng-Shuai and Liu, Zhijian and Liu, Xihui and Qiao, Yu and Ouyang, Wanli and He, Tong and Zhao, Hengshuang},
    booktitle={CVPR},
    year={2024}
}

@inproceedings{masked_scene_contrast,
  author    = {Wu, Xiaoyang and Wen, Xin and Liu, Xihui and Zhao, Hengshuang},
  title     = {Masked Scene Contrast: A Scalable Framework for Unsupervised {3D} Representation Learning},
  booktitle = {Proceedings of the IEEE/CVF Conference on Computer Vision and Pattern Recognition (CVPR)},
  pages     = {9415--9424},
  year      = {2023}
}

@misc{pointcept,
    title={Pointcept: A Codebase for Point Cloud Perception Research},
    author={Pointcept Contributors},
    howpublished={\url{https://github.com/Pointcept/Pointcept}},
    year={2023}
}

@inproceedings{ptv2,
    title={Point transformer V2: Grouped Vector Attention and Partition-based Pooling},
    author={Wu, Xiaoyang and Lao, Yixing and Jiang, Li and Liu, Xihui and Zhao, Hengshuang},
    booktitle={NeurIPS},
    year={2022}
}

@INPROCEEDINGS{ptv1,
  author={Zhao, Hengshuang and Jiang, Li and Jia, Jiaya and Torr, Philip and Koltun, Vladlen},
  booktitle={2021 IEEE/CVF International Conference on Computer Vision (ICCV)}, 
  title={Point Transformer}, 
  year={2021},
  volume={},
  number={},
  pages={16239-16248},
  doi={10.1109/ICCV48922.2021.01595}
}

@inproceedings{moco,
  author={He, Kaiming and Fan, Haoqi and Wu, Yuxin and Xie, Saining and Girshick, Ross},
  booktitle={2020 IEEE/CVF Conference on Computer Vision and Pattern Recognition (CVPR)}, 
  title={Momentum Contrast for Unsupervised Visual Representation Learning}, 
  year={2020},
  volume={},
  number={},
  pages={9726-9735},
  doi={10.1109/CVPR42600.2020.00975}
}

@inproceedings{sal,
author = {O{\v s}ep, Aljo{\v s}a and Meinhardt, Tim and Ferroni, Francesco and Peri, Neehar and Ramanan, Deva and Leal-Taix{\'e}, Laura},
title = {Better Call SAL: Towards Learning to Segment Anything in Lidar},
year = {2024},
url = {https://doi.org/10.1007/978-3-031-72933-1_5},
doi = {10.1007/978-3-031-72933-1_5},
booktitle = {ECCV 2024},
pages = {71–90},
numpages = {20}
}

@inproceedings{bosch_nomae,
  author    = {Abdelsamad, Mohamed and Ulrich, Michael and Gl{\"a}ser, Claudius and Valada, Abhinav},
  title     = {Multi-Scale Neighborhood Occupancy Masked Autoencoder for Self-Supervised Learning in {LiDAR} Point Clouds},
  booktitle = {Proceedings of the IEEE/CVF Conference on Computer Vision and Pattern Recognition (CVPR)},
  pages     = {22234--22243},
  year      = {2025},
  doi       = {10.1109/CVPR52734.2025.02071}
}

@inproceedings{uno,
  author={Agro, Ben and Sykora, Quinlan and Casas, Sergio and Gilles, Thomas and Urtasun, Raquel},
  booktitle={2024 IEEE/CVF Conference on Computer Vision and Pattern Recognition (CVPR)}, 
  title={UnO: Unsupervised Occupancy Fields for Perception and Forecasting}, 
  year={2024},
  volume={},
  number={},
  pages={14487-14496},
  doi={10.1109/CVPR52733.2024.01373}
}

@inproceedings{4docc,
author = { Khurana, Tarasha and Hu, Peiyun and Held, David and Ramanan, Deva },
booktitle = { 2023 IEEE/CVF Conference on Computer Vision and Pattern Recognition (CVPR) },
title = {{ Point Cloud Forecasting as a Proxy for 4D Occupancy Forecasting }},
year = {2023},
volume = {},
ISSN = {},
pages = {1116-1124},
doi = {10.1109/CVPR52729.2023.00114},
url = {https://doi.ieeecomputersociety.org/10.1109/CVPR52729.2023.00114},
publisher = {IEEE Computer Society},
month =Jun
}

@article{gasp,
  title        = {GASP: Unifying Geometric and Semantic Self-Supervised Pre-training for Autonomous Driving},
  author       = {Ljungbergh, William and Lilja, Adam and Tonderski, Adam and Laveno Ling, Arvid and Lindstr{\"o}m, Carl and Verbeke, Willem and Fu, Junsheng and Petersson, Christoffer and Hammarstrand, Lars and Felsberg, Michael},
  journal      = {WACV},
  year         = {2026}
}

@inproceedings{simclr,
  author    = {Chen, Ting and Kornblith, Simon and Norouzi, Mohammad and Hinton, Geoffrey},
  title     = {A Simple Framework for Contrastive Learning of Visual Representations},
  booktitle = {International Conference on Machine Learning (ICML)},
  series    = {Proceedings of Machine Learning Research},
  volume    = {119},
  pages     = {1597--1607},
  year      = {2020}
}

@article{rope,
title = {RoFormer: Enhanced transformer with Rotary Position Embedding},
journal = {Neurocomputing},
volume = {568},
pages = {127063},
year = {2024},
issn = {0925-2312},
doi = {https://doi.org/10.1016/j.neucom.2023.127063},
url = {https://www.sciencedirect.com/science/article/pii/S0925231223011864},
author = {Jianlin Su and Murtadha Ahmed and Yu Lu and Shengfeng Pan and Wen Bo and Yunfeng Liu},
}

@inproceedings{litept,
    title={{LitePT: Lighter Yet Stronger Point Transformer}},
    author={Yue, Yuanwen and Robert, Damien and Wang, Jianyuan and Hong, Sunghwan and Wegner, Jan Dirk and Rupprecht, Christian and Schindler, Konrad},
    booktitle={IEEE/CVF Conference on Computer Vision and Pattern Recognition (CVPR)},
    year={2026}
}

@article{deformable_detr,
  title={Deformable DETR: Deformable Transformers for End-to-End Object Detection},
  author={Zhu, Xizhou and Su, Weijie and Lu, Lewei and Li, Bin and Wang, Xiaogang and Dai, Jifeng},
  journal={arXiv preprint arXiv:2010.04159},
  year={2020}
}

@inproceedings{vicreg,
title={{VICR}eg: Variance-Invariance-Covariance Regularization for Self-Supervised Learning},
author={Adrien Bardes and Jean Ponce and Yann LeCun},
booktitle={International Conference on Learning Representations},
year={2022},
url={https://openreview.net/forum?id=xm6YD62D1Ub}
}

@misc{sigreg,
      title={LeJEPA: Provable and Scalable Self-Supervised Learning Without the Heuristics}, 
      author={Randall Balestriero and Yann LeCun},
      year={2025},
      eprint={2511.08544},
      archivePrefix={arXiv},
      primaryClass={cs.LG},
      url={https://arxiv.org/abs/2511.08544}, 
}

@inproceedings{barlow_twins,
  title = 	 {Barlow Twins: Self-Supervised Learning via Redundancy Reduction},
  author =       {Zbontar, Jure and Jing, Li and Misra, Ishan and LeCun, Yann and Deny, Stephane},
  booktitle = 	 {Proceedings of the 38th International Conference on Machine Learning},
  pages = 	 {12310--12320},
  year = 	 {2021},
  editor = 	 {Meila, Marina and Zhang, Tong},
  volume = 	 {139},
  series = 	 {Proceedings of Machine Learning Research},
  month = 	 {18--24 Jul},
  publisher =    {PMLR},
  url = 	 {https://proceedings.mlr.press/v139/zbontar21a.html},
}

@inproceedings{byol,
  author    = {Grill, Jean-Bastien and Strub, Florian and Altch{\'e}, Florent and Tallec, Corentin and Richemond, Pierre H. and Buchatskaya, Elena and Doersch, Carl and Pires, Bernardo Avila and Guo, Zhaohan Daniel and Azar, Mohammad Gheshlaghi and Piot, Bilal and Kavukcuoglu, Koray and Munos, R{\'e}mi and Valko, Michal},
  title     = {Bootstrap Your Own Latent: A New Approach to Self-Supervised Learning},
  booktitle = {Advances in Neural Information Processing Systems (NeurIPS)},
  volume    = {33},
  pages     = {21271--21284},
  year      = {2020}
}

@INPROCEEDINGS{dino,
  author={Caron, Mathilde and Touvron, Hugo and Misra, Ishan and Jegou, Hervé and Mairal, Julien and Bojanowski, Piotr and Joulin, Armand},
  booktitle={2021 IEEE/CVF International Conference on Computer Vision (ICCV)}, 
  title={Emerging Properties in Self-Supervised Vision Transformers}, 
  year={2021},
  volume={},
  number={},
  pages={9630-9640},
  doi={10.1109/ICCV48922.2021.00951}
}

@article{dinov2,
title={{DINO}v2: Learning Robust Visual Features without Supervision},
author={Maxime Oquab and Timoth{\'e}e Darcet and Th{\'e}o Moutakanni and Huy V. Vo and Marc Szafraniec and Vasil Khalidov and Pierre Fernandez and Daniel HAZIZA and Francisco Massa and Alaaeldin El-Nouby and Mido Assran and Nicolas Ballas and Wojciech Galuba and Russell Howes and Po-Yao Huang and Shang-Wen Li and Ishan Misra and Michael Rabbat and Vasu Sharma and Gabriel Synnaeve and Hu Xu and Herve Jegou and Julien Mairal and Patrick Labatut and Armand Joulin and Piotr Bojanowski},
journal={Transactions on Machine Learning Research},
issn={2835-8856},
year={2024},
url={https://openreview.net/forum?id=a68SUt6zFt},
note={Featured Certification}
}

@misc{dinov3,
      title={DINOv3}, 
      author={Oriane Siméoni and Huy V. Vo and Maximilian Seitzer and Federico Baldassarre and Maxime Oquab and Cijo Jose and Vasil Khalidov and Marc Szafraniec and Seungeun Yi and Michaël Ramamonjisoa and Francisco Massa and Daniel Haziza and Luca Wehrstedt and Jianyuan Wang and Timothée Darcet and Théo Moutakanni and Leonel Sentana and Claire Roberts and Andrea Vedaldi and Jamie Tolan and John Brandt and Camille Couprie and Julien Mairal and Hervé Jégou and Patrick Labatut and Piotr Bojanowski},
      year={2025},
      eprint={2508.10104},
      archivePrefix={arXiv},
      primaryClass={cs.CV},
      url={https://arxiv.org/abs/2508.10104}, 
}

@misc{a_path_towards,
  author       = {LeCun, Yann},
  title        = {A Path Towards Autonomous Machine Intelligence},
  howpublished = {OpenReview preprint},
  year         = {2022},
  url          = {https://openreview.net/forum?id=BZ5a1r-kVsf}
}

@article{ijepa,
  title={Self-Supervised Learning from Images with a Joint-Embedding Predictive Architecture},
  author={Assran, Mahmoud and Duval, Quentin and Misra, Ishan and Bojanowski, Piotr and Vincent, Pascal and Rabbat, Michael and LeCun, Yann and Ballas, Nicolas},
  journal={arXiv preprint arXiv:2301.08243},
  year={2023}
}

@inproceedings{concerto,
  title={Concerto: Joint 2D-3D Self-Supervised Learning Emerges Spatial Representations},
  author={Zhang, Yujia and Wu, Xiaoyang and Lao, Yixing and Wang, Chengyao and Tian, Zhuotao and Wang, Naiyan and Zhao, Hengshuang},
  booktitle={NeurIPS},
  year={2025}
}

@inproceedings{bosch_dos,
author = {Abdelsamad, Mohamed and Ulrich, Michael and Yang, Bin and Zhang, Miao and Miron, Yakov and Valada, Abhinav},
title = {DOS: distilling observable softmaps of zipfian prototypes for self-supervised point representation},
year = {2026},
isbn = {978-1-57735-906-7},
publisher = {AAAI Press},
url = {https://doi.org/10.1609/aaai.v40i24.39030},
doi = {10.1609/aaai.v40i24.39030},
booktitle = {Proceedings of the Fortieth AAAI Conference on Artificial Intelligence and Thirty-Eighth Conference on Innovative Applications of Artificial Intelligence and Sixteenth Symposium on Educational Advances in Artificial Intelligence},
articleno = {2174},
numpages = {10},
series = {AAAI'26/IAAI'26/EAAI'26}
}

@inproceedings{vit,
title={An Image is Worth 16x16 Words: Transformers for Image Recognition at Scale},
author={Alexey Dosovitskiy and Lucas Beyer and Alexander Kolesnikov and Dirk Weissenborn and Xiaohua Zhai and Thomas Unterthiner and Mostafa Dehghani and Matthias Minderer and Georg Heigold and Sylvain Gelly and Jakob Uszkoreit and Neil Houlsby},
booktitle={International Conference on Learning Representations},
year={2021},
url={https://openreview.net/forum?id=YicbFdNTTy}
}

@article{param_hep,
   title={Parameterized neural networks for high-energy physics},
   volume={76},
   ISSN={1434-6052},
   url={http://dx.doi.org/10.1140/epjc/s10052-016-4099-4},
   DOI={10.1140/epjc/s10052-016-4099-4},
   number={5},
   journal={The European Physical Journal C},
   publisher={Springer Science and Business Media LLC},
   author={Baldi, Pierre and Cranmer, Kyle and Faucett, Taylor and Sadowski, Peter and Whiteson, Daniel},
   year={2016},
   month=Apr
}

@article{hep_ewkino,
   title={Search for electroweak production of charginos and neutralinos in proton-proton collisions at $\sqrt{s} = 13$ $\text{TeV}$},
   volume={2022},
   ISSN={1029-8479},
   url={http://dx.doi.org/10.1007/JHEP04(2022)147},
   DOI={10.1007/jhep04(2022)147},
   number={4},
   journal={Journal of High Energy Physics},
   publisher={Springer Science and Business Media LLC},
   author={CMS Collaboration},
   year={2022},
   month=Apr
}

@article{vjepa,
title={Revisiting Feature Prediction for Learning Visual Representations from Video},
author={Adrien Bardes and Quentin Garrido and Jean Ponce and Xinlei Chen and Michael Rabbat and Yann LeCun and Mido Assran and Nicolas Ballas},
journal={Transactions on Machine Learning Research},
issn={2835-8856},
year={2024},
url={https://openreview.net/forum?id=QaCCuDfBk2},
note={Featured Certification}
}

@article{vjepa_2,
  author  = {Assran, Mido and Bardes, Adrien and Fan, David and Garrido, Quentin and Howes, Russell and Komeili, Mojtaba and Muckley, Matthew and Rizvi, Ammar and Roberts, Claire and Sinha, Koustuv and Zholus, Artem and Arnaud, Sergio and Gejji, Abha and Martin, Ada and Hogan, Francois Robert and Dugas, Daniel and Bojanowski, Piotr and Khalidov, Vasil and Labatut, Patrick and Massa, Francisco and Szafraniec, Marc and Krishnakumar, Kapil and Li, Yong and Ma, Xiaodong and Chandar, Sarath and Meier, Franziska and LeCun, Yann and Rabbat, Michael and Ballas, Nicolas},
  title   = {{V-JEPA 2}: Self-Supervised Video Models Enable Understanding, Prediction and Planning},
  journal = {arXiv preprint arXiv:2506.09985},
  year    = {2025}
}

@article{vjepa_2p1,
  publtype={informal},
  author={Lorenzo Mur-Labadia and Matthew J. Muckley and Amir Bar and Mido Assran and Koustuv Sinha and Mike Rabbat and Yann LeCun and Nicolas Ballas and Adrien Bardes},
  title={V-JEPA 2.1: Unlocking Dense Features in Video Self-Supervised Learning},
  year={2026},
  month={March},
  cdate={1772323200000},
  journal={CoRR},
  volume={abs/2603.14482},
  url={https://doi.org/10.48550/arXiv.2603.14482}
}

@inproceedings{bosch_pointins,
title={Towards Foundation Models for 3D Scene Understanding: Instance-Aware Self-Supervised Learning for Point Clouds},
author={Bin Yang and Mohamed Abdelsamad and Miao Zhang and Alexandru Paul Condurache},
booktitle={Conference on Computer Vision and Pattern Recognition 2026},
year={2026},
url={https://openreview.net/forum?id=8ADGzRtQDt}
}

@inproceedings{bosch_ghostpoint,
  author    = {Abdelsamad, Mohamed and Yang, Bin and Ulrich, Michael and Zhang, Miao and Miron, Yakov and Condurache, Alexandru Paul and Valada, Abhinav},
  title     = {{GhostPoint}: Self-Supervised Representation Learning by Hallucinating Occluded {LiDAR} Structure},
  booktitle = {European Conference on Computer Vision (ECCV)},
  pages     = {116--137},
  year      = {2026},
  doi       = {10.1007/978-3-032-37369-4_7}
}

@InProceedings{dit,
    author    = {Peebles, William and Xie, Saining},
    title     = {Scalable Diffusion Models with Transformers},
    booktitle = {Proceedings of the IEEE/CVF International Conference on Computer Vision (ICCV)},
    month     = {October},
    year      = {2023},
    pages     = {4195-4205}
}

@inproceedings{film,
author = {Perez, Ethan and Strub, Florian and de Vries, Harm and Dumoulin, Vincent and Courville, Aaron},
title = {FiLM: visual reasoning with a general conditioning layer},
year = {2018},
isbn = {978-1-57735-800-8},
publisher = {AAAI Press},
booktitle = {Proceedings of the Thirty-Second AAAI Conference on Artificial Intelligence and Thirtieth Innovative Applications of Artificial Intelligence Conference and Eighth AAAI Symposium on Educational Advances in Artificial Intelligence},
articleno = {483},
numpages = {10},
location = {New Orleans, Louisiana, USA},
series = {AAAI'18/IAAI'18/EAAI'18}
}

@inproceedings{lidar_clip,
  author={Hess, Georg and Tonderski, Adam and Petersson, Christoffer and Åström, Kalle and Svensson, Lennart},
  booktitle={2024 IEEE/CVF Winter Conference on Applications of Computer Vision (WACV)}, 
  title={LidarCLIP or: How I Learned to Talk to Point Clouds}, 
  year={2024},
  volume={},
  number={},
  pages={7423-7432},
  doi={10.1109/WACV57701.2024.00727}
  }

@inproceedings{georg_pointmae,
   title={Masked Autoencoder for Self-Supervised Pre-training on Lidar Point Clouds},
   url={http://dx.doi.org/10.1109/WACVW58289.2023.00039},
   DOI={10.1109/wacvw58289.2023.00039},
   booktitle={2023 IEEE/CVF Winter Conference on Applications of Computer Vision Workshops (WACVW)},
   publisher={IEEE},
   author={Hess, Georg and Jaxing, Johan and Svensson, Elias and Hagerman, David and Petersson, Christoffer and Svensson, Lennart},
   year={2023},
   month=Jan, pages={350–359}
}

@inproceedings{pointmae,
  title={Masked autoencoders for point cloud self-supervised learning},
  author={Pang, Yatian and Wang, Wenxiao and Tay, Francis EH and Liu, Wei and Tian, Yonghong and Yuan, Li},
  booktitle={Computer Vision--ECCV 2022: 17th European Conference, Tel Aviv, Israel, October 23--27, 2022, Proceedings, Part II},
  pages={604--621},
  year={2022},
  organization={Springer}
}

@inproceedings{sdpa_gate,
    title={Gated Attention for Large Language Models: Non-linearity, Sparsity, and Attention-Sink-Free},
    author={Zihan Qiu and Zekun Wang and Bo Zheng and Zeyu Huang and Kaiyue Wen and Songlin Yang and Rui Men and Le Yu and Fei Huang and Suozhi Huang and Dayiheng Liu and Jingren Zhou and Junyang Lin},
    booktitle={The Thirty-ninth Annual Conference on Neural Information Processing Systems},
    year={2025},
    url={https://openreview.net/forum?id=1b7whO4SfY}
}

@inproceedings{mae,
  author    = {He, Kaiming and Chen, Xinlei and Xie, Saining and Li, Yanghao and Doll{\'a}r, Piotr and Girshick, Ross},
  title     = {Masked Autoencoders Are Scalable Vision Learners},
  booktitle = {Proceedings of the IEEE/CVF Conference on Computer Vision and Pattern Recognition (CVPR)},
  pages     = {15979--15988},
  year      = {2022},
  doi       = {10.1109/CVPR52688.2022.01553}
}

@misc{ad_jepa_wm,
      title={Self-Supervised JEPA-based World Models for LiDAR Occupancy Completion and Forecasting}, 
      author={Haoran Zhu and Anna Choromanska},
      year={2026},
      eprint={2602.12540},
      archivePrefix={arXiv},
      primaryClass={cs.CV},
      url={https://arxiv.org/abs/2602.12540}, 
}

@inproceedings{hilda,
  title     = {HilDA: Hierarchical Distillation with Diffusion for
               Advancing Self-Supervised LiDAR Pre-training},
  author    = {Wozniak, Maciej and Ericsson, Jesper and
               Govindarajan, Hariprasath and Nyberg, Truls and
               Gustafsson, Thomas and Jensfelt, Patric and Andersson, Olov},
  booktitle = {European Conference on Computer Vision (ECCV)},
  year      = {2026}
}

@inproceedings{volt,
  title     = {{Volume Transformer: Revisiting Vanilla Transformers for 3D Scene Understanding}},
  author    = {Yilmaz, Kadir and Kruse, Adrian and H{\"o}fer, Tristan and de Geus, Daan and Leibe, Bastian},
  booktitle = {European Conference on Computer Vision (ECCV)},
  year      = {2026},
}

@INPROCEEDINGS {argoverse2,
  author = {Benjamin Wilson and William Qi and Tanmay Agarwal and John Lambert and Jagjeet Singh and Siddhesh Khandelwal and Bowen Pan and Ratnesh Kumar and Andrew Hartnett and Jhony Kaesemodel Pontes and Deva Ramanan and Peter Carr and James Hays},
  title = {Argoverse 2: Next Generation Datasets for Self-Driving Perception and Forecasting},
  booktitle = {Proceedings of the Neural Information Processing Systems Track on Datasets and Benchmarks (NeurIPS Datasets and Benchmarks 2021)},
  year = {2021}
}

@inproceedings{ad_l_jepa,
  author    = {Zhu, Haoran and Dong, Zhenyuan and Topollai, Kristi and Sha, Beiyao and Choromanska, Anna Ewa},
  title     = {Self-Supervised Representation Learning with Joint Embedding Predictive Architecture for Automotive {LiDAR} Object Detection},
  booktitle = {Proceedings of the AAAI Conference on Artificial Intelligence},
  volume    = {40},
  number    = {16},
  year      = {2026},
  doi       = {10.1609/aaai.v40i16.38402}
}

@INPROCEEDINGS{point_jepa,
  author={Saito, Ayumu and Kudeshia, Prachi and Poovvancheri, Jiju},
  booktitle={2025 IEEE/CVF Winter Conference on Applications of Computer Vision (WACV)}, 
  title={Point-JEPA: A Joint Embedding Predictive Architecture for Self-Supervised Learning on Point Cloud}, 
  year={2025},
  volume={},
  number={},
  pages={7348-7357},
  doi={10.1109/WACV61041.2025.00714}
}

@techreport{gpt2,
  author      = {Radford, Alec and Wu, Jeff and Child, Rewon and Luan, David and Amodei, Dario and Sutskever, Ilya},
  title       = {Language Models are Unsupervised Multitask Learners},
  institution = {OpenAI},
  year        = {2019}
}

@misc{llama,
      title={LLaMA: Open and Efficient Foundation Language Models}, 
      author={Hugo Touvron and Thibaut Lavril and Gautier Izacard and Xavier Martinet and Marie-Anne Lachaux and Timothée Lacroix and Baptiste Rozière and Naman Goyal and Eric Hambro and Faisal Azhar and Aurelien Rodriguez and Armand Joulin and Edouard Grave and Guillaume Lample},
      year={2023},
      eprint={2302.13971},
      archivePrefix={arXiv},
      primaryClass={cs.CL},
      url={https://arxiv.org/abs/2302.13971}, 
}

@misc{deepseekllm,
      title={DeepSeek LLM: Scaling Open-Source Language Models with Longtermism}, 
      author={DeepSeek-AI and Xiao Bi and Deli Chen and Guanting Chen and Shanhuang Chen and Damai Dai and Chengqi Deng and Honghui Ding and Kai Dong and Qiushi Du and Zhe Fu and Huazuo Gao and Kaige Gao and Wenjun Gao and Ruiqi Ge and Kang Guan and Daya Guo and Jianzhong Guo and Guangbo Hao and Zhewen Hao and Ying He and Wenjie Hu and Panpan Huang and Erhang Li and Guowei Li and Jiashi Li and Yao Li and Y. K. Li and Wenfeng Liang and Fangyun Lin and A. X. Liu and Bo Liu and Wen Liu and Xiaodong Liu and Xin Liu and Yiyuan Liu and Haoyu Lu and Shanghao Lu and Fuli Luo and Shirong Ma and Xiaotao Nie and Tian Pei and Yishi Piao and Junjie Qiu and Hui Qu and Tongzheng Ren and Zehui Ren and Chong Ruan and Zhangli Sha and Zhihong Shao and Junxiao Song and Xuecheng Su and Jingxiang Sun and Yaofeng Sun and Minghui Tang and Bingxuan Wang and Peiyi Wang and Shiyu Wang and Yaohui Wang and Yongji Wang and Tong Wu and Y. Wu and Xin Xie and Zhenda Xie and Ziwei Xie and Yiliang Xiong and Hanwei Xu and R. X. Xu and Yanhong Xu and Dejian Yang and Yuxiang You and Shuiping Yu and Xingkai Yu and B. Zhang and Haowei Zhang and Lecong Zhang and Liyue Zhang and Mingchuan Zhang and Minghua Zhang and Wentao Zhang and Yichao Zhang and Chenggang Zhao and Yao Zhao and Shangyan Zhou and Shunfeng Zhou and Qihao Zhu and Yuheng Zou},
      year={2024},
      eprint={2401.02954},
      archivePrefix={arXiv},
      primaryClass={cs.CL},
      url={https://arxiv.org/abs/2401.02954}, 
}

@misc{ad_jepa_planning,
      title={AD-E2E-JEPA: A Joint-Embedding Predictive Architecture For End-to-End Autonomous Driving}, 
      author={Haoran Zhu and Wancong Zhang and Yann LeCun and Anna Choromanska},
      year={2026},
      eprint={2609.34085},
      archivePrefix={arXiv},
      primaryClass={cs.RO},
      url={https://arxiv.org/abs/2609.34085}, 
}

@inproceedings{dino_wm,
    title={{DINO}-{WM}: World Models on Pre-trained Visual Features enable Zero-shot Planning},
    author={Gaoyue Zhou and Hengkai Pan and Yann LeCun and Lerrel Pinto},
    booktitle={Forty-second International Conference on Machine Learning},
    year={2025},
    url={https://openreview.net/forum?id=D5RNACOZEI}
}

@inproceedings{vggt_omega,
  title     = {{VGGT-$\Omega$}},
  author    = {Jianyuan Wang and Minghao Chen and Shangzhan Zhang and Nikita Karaev and Johannes Sch{\"o}nberger and Patrick Labatut and Piotr Bojanowski and David Novotny and Andrea Vedaldi and Christian Rupprecht},
  booktitle = {Proceedings of the IEEE/CVF Conference on Computer Vision and Pattern Recognition (CVPR)},
  year      = {2026}
}

@article{submanifold_convs,
  title={Submanifold Sparse Convolutional Networks},
  author={Graham, Benjamin and van der Maaten, Laurens},
  journal={arXiv preprint arXiv:1706.01307},
  year={2017}
}

@misc{adamw,
      title={Decoupled Weight Decay Regularization}, 
      author={Ilya Loshchilov and Frank Hutter},
      year={2019},
      eprint={1711.05101},
      archivePrefix={arXiv},
      primaryClass={cs.LG},
      url={https://arxiv.org/abs/1711.05101}, 
}

@inproceedings{vit_22b,
  title = 	 {Scaling Vision Transformers to 22 Billion Parameters},
  author =       {Dehghani, Mostafa and Djolonga, Josip and Mustafa, Basil and Padlewski, Piotr and Heek, Jonathan and Gilmer, Justin and Steiner, Andreas Peter and Caron, Mathilde and Geirhos, Robert and Alabdulmohsin, Ibrahim and Jenatton, Rodolphe and Beyer, Lucas and Tschannen, Michael and Arnab, Anurag and Wang, Xiao and Riquelme Ruiz, Carlos and Minderer, Matthias and Puigcerver, Joan and Evci, Utku and Kumar, Manoj and Steenkiste, Sjoerd Van and Elsayed, Gamaleldin Fathy and Mahendran, Aravindh and Yu, Fisher and Oliver, Avital and Huot, Fantine and Bastings, Jasmijn and Collier, Mark and Gritsenko, Alexey A. and Birodkar, Vighnesh and Vasconcelos, Cristina Nader and Tay, Yi and Mensink, Thomas and Kolesnikov, Alexander and Pavetic, Filip and Tran, Dustin and Kipf, Thomas and Lucic, Mario and Zhai, Xiaohua and Keysers, Daniel and Harmsen, Jeremiah J. and Houlsby, Neil},
  booktitle = 	 {Proceedings of the 40th International Conference on Machine Learning},
  pages = 	 {7480--7512},
  year = 	 {2023},
  editor = 	 {Krause, Andreas and Brunskill, Emma and Cho, Kyunghyun and Engelhardt, Barbara and Sabato, Sivan and Scarlett, Jonathan},
  volume = 	 {202},
  series = 	 {Proceedings of Machine Learning Research},
  month = 	 {23--29 Jul},
  publisher =    {PMLR},
  url = 	 {https://proceedings.mlr.press/v202/dehghani23a.html},
}

@misc{zod,
      title={Zenseact Open Dataset: A large-scale and diverse multimodal dataset for autonomous driving}, 
      author={Mina Alibeigi and William Ljungbergh and Adam Tonderski and Georg Hess and Adam Lilja and Carl Lindstrom and Daria Motorniuk and Junsheng Fu and Jenny Widahl and Christoffer Petersson},
      year={2023},
      eprint={2305.02008},
      archivePrefix={arXiv},
      primaryClass={cs.CV},
      url={https://arxiv.org/abs/2305.02008}, 
}

@inproceedings{adaln_single,
  title     = {PixArt-$\alpha$: Fast Training of Diffusion Transformer for Photorealistic Text-to-Image Synthesis},
  author    = {Chen, Junsong and Yu, Jincheng and Ge, Chongjian and Yao, Lewei and Xie, Enze and Wang, Zhongdao and Kwok, James and Luo, Ping and Lu, Huchuan and Li, Zhenguo},
  booktitle = {International Conference on Learning Representations (ICLR)},
  year      = {2024}
}

@inproceedings{waymo,
    author = {Sun, Pei and Kretzschmar, Henrik and Dotiwalla, Xerxes and Chouard, Aurelien and Patnaik, Vijaysai and Tsui, Paul and Guo, James and Zhou, Yin and Chai, Yuning and Caine, Benjamin and Vasudevan, Vijay and Han, Wei and Ngiam, Jiquan and Zhao, Hang and Timofeev, Aleksei and Ettinger, Scott and Krivokon, Maxim and Gao, Amy and Joshi, Aditya and Zhang, Yu and Shlens, Jonathon and Chen, Zhifeng and Anguelov, Dragomir},
    title = {Scalability in Perception for Autonomous Driving: Waymo Open Dataset},
    booktitle = {Proceedings of the IEEE/CVF Conference on Computer Vision and Pattern Recognition (CVPR)},
    month = {June},
    year = {2020}
}

@article{pointgroup,
  title={PointGroup: Dual-Set Point Grouping for 3D Instance Segmentation},
  author={Jiang, Li and Zhao, Hengshuang and Shi, Shaoshuai and Liu, Shu and Fu, Chi-Wing and Jia, Jiaya},
  journal={Proceedings of the IEEE Conference on Computer Vision and Pattern Recognition (CVPR)},
  year={2020}
}

@misc{unet,
      title={U-Net: Convolutional Networks for Biomedical Image Segmentation}, 
      author={Olaf Ronneberger and Philipp Fischer and Thomas Brox},
      year={2015},
      eprint={1505.04597},
      archivePrefix={arXiv},
      primaryClass={cs.CV},
      url={https://arxiv.org/abs/1505.04597}, 
}

@misc{nuscenes,
      title={nuScenes: A multimodal dataset for autonomous driving}, 
      author={Holger Caesar and Varun Bankiti and Alex H. Lang and Sourabh Vora and Venice Erin Liong and Qiang Xu and Anush Krishnan and Yu Pan and Giancarlo Baldan and Oscar Beijbom},
      year={2020},
      eprint={1903.11027},
      archivePrefix={arXiv},
      primaryClass={cs.LG},
      url={https://arxiv.org/abs/1903.11027}, 
}

@misc{cirrus,
      title={Cirrus: A Long-range Bi-pattern LiDAR Dataset}, 
      author={Ze Wang and Sihao Ding and Ying Li and Jonas Fenn and Sohini Roychowdhury and Andreas Wallin and Lane Martin and Scott Ryvola and Guillermo Sapiro and Qiang Qiu},
      year={2020},
      eprint={2012.02938},
      archivePrefix={arXiv},
      primaryClass={cs.CV},
      url={https://arxiv.org/abs/2012.02938}, 
}

@misc{truckscenes,
      title={MAN TruckScenes: A multimodal dataset for autonomous trucking in diverse conditions}, 
      author={Felix Fent and Fabian Kuttenreich and Florian Ruch and Farija Rizwin and Stefan Juergens and Lorenz Lechermann and Christian Nissler and Andrea Perl and Ulrich Voll and Min Yan and Markus Lienkamp},
      year={2024},
      eprint={2407.07462},
      archivePrefix={arXiv},
      primaryClass={cs.CV},
      url={https://arxiv.org/abs/2407.07462}, 
}

@misc{octosense,
      title={OctoSense: Self-Supervised Learning for Multimodal Robot Perception}, 
      author={Anthony Bisulco and Jeremy Wang and Kostas Daniilidis and Randall Balestriero and Pratik Chaudhari},
      year={2026},
      eprint={2606.27317},
      archivePrefix={arXiv},
      primaryClass={cs.CV},
      url={https://arxiv.org/abs/2606.27317}, 
}

@INPROCEEDINGS{pandaset,
  author={Xiao, Pengchuan and Shao, Zhenlei and Hao, Steven and Zhang, Zishuo and Chai, Xiaolin and Jiao, Judy and Li, Zesong and Wu, Jian and Sun, Kai and Jiang, Kun and Wang, Yunlong and Yang, Diange},
  booktitle={2021 IEEE International Intelligent Transportation Systems Conference (ITSC)}, 
  title={PandaSet: Advanced Sensor Suite Dataset for Autonomous Driving}, 
  year={2021},
  pages={3095-3101},
  doi={10.1109/ITSC48978.2021.9565009}}

@inproceedings{semantic_kitti_base,
  author = {A. Geiger and P. Lenz and R. Urtasun},
  title = {{Are we ready for Autonomous Driving? The KITTI Vision Benchmark Suite}},
  booktitle = {Proc.~of the IEEE Conf.~on Computer Vision and Pattern Recognition (CVPR)},
  pages = {3354--3361},
  year = {2012}
}

@inproceedings{semantic_kitti,
  author = {J. Behley and M. Garbade and A. Milioto and J. Quenzel and S. Behnke and C. Stachniss and J. Gall},
  title = {{SemanticKITTI: A Dataset for Semantic Scene Understanding of LiDAR Sequences}},
  booktitle = {Proc. of the IEEE/CVF International Conf.~on Computer Vision (ICCV)},
  year = {2019}
}

@article{kitti_raw,
  author = {Andreas Geiger and Philip Lenz and Christoph Stiller and Raquel Urtasun},
  title = {Vision meets Robotics: The KITTI Dataset},
  journal = {International Journal of Robotics Research (IJRR)},
  year = {2013}
}

@inproceedings{depth_anything_v2,
    title={Depth Anything V2},
    author={Lihe Yang and Bingyi Kang and Zilong Huang and Zhen Zhao and Xiaogang Xu and Jiashi Feng and Hengshuang Zhao},
    booktitle={The Thirty-eighth Annual Conference on Neural Information Processing Systems},
    year={2024},
    url={https://openreview.net/forum?id=cFTi3gLJ1X}
}

@inproceedings{lvsm,
    title={{LVSM}: A Large View Synthesis Model with Minimal 3D Inductive Bias},
    author={Haian Jin and Hanwen Jiang and Hao Tan and Kai Zhang and Sai Bi and Tianyuan Zhang and Fujun Luan and Noah Snavely and Zexiang Xu},
    booktitle={The Thirteenth International Conference on Learning Representations},
    year={2025},
    url={https://openreview.net/forum?id=QQBPWtvtcn}
}

@InProceedings{vggt,
    author    = {Wang, Jianyuan and Chen, Minghao and Karaev, Nikita and Vedaldi, Andrea and Rupprecht, Christian and Novotny, David},
    title     = {VGGT: Visual Geometry Grounded Transformer},
    booktitle = {Proceedings of the IEEE/CVF Conference on Computer Vision and Pattern Recognition (CVPR)},
    month     = {June},
    year      = {2025},
    pages     = {5294-5306}
}

@inproceedings{asymmetric_self_distill_point,
    title={Asymmetric Dual Self-Distillation for 3D Self-Supervised Representation Learning},
    author={Remco F. Leijenaar and Hamidreza Kasaei},
    booktitle={The Thirty-ninth Annual Conference on Neural Information Processing Systems},
    year={2025},
    url={https://openreview.net/forum?id=ANl9HuwIMs}
}

@article{hilbert_curve,
  author  = {Hilbert, David},
  title   = {Ueber die stetige {A}bbildung einer {L}inie auf ein {F}l{\"a}chenst{\"u}ck},
  journal = {Mathematische Annalen},
  volume  = {38},
  number  = {3},
  pages   = {459--460},
  year    = {1891},
  url     = {https://eudml.org/doc/157555}
}

@book{zorder,
  author    = {Lebesgue, Henri},
  title     = {Le{\c{c}}ons sur l'int{\'e}gration et la recherche des fonctions primitives, profess{\'e}es au Coll{\`e}ge de France},
  publisher = {Gauthier-Villars},
  address   = {Paris},
  year      = {1904},
  pages     = {44--45}
}
}

\clearpage
\maketitlesupplementary

\section{Extended Related Work} \label{sup:extended_related_work}

The development of \ac{SSL} methods for 3D point clouds is closely related to \ac{SSL} for images and video. Techniques can be broadly distinguished by whether they try to directly reconstruct unseen data, distill existing models, or learn in their own embedding space.

\parsection{Generative and semi-supervised methods}
Generative methods reconstruct corrupted or partially observed data and have been applied to point clouds~\cite{georg_pointmae, pointmae, bosch_nomae}. Without fine-tuning for new downstream tasks, this approach often results in weak performance~\cite{mae}. We aim for a single encoder ready for use on a large array of tasks, in which case generative methods are not the pragmatic choice.

Other methods derive heuristics, often occupancy or flow~\cite{4docc, uno}, that the model predicts during training. This focuses on relevant information without having to learn arbitrarily complex physics or sensor intrinsics. Point cloud encoders can also be trained by distilling information from vision foundation models~\cite{sal, lidar_clip, hilda}. Both approaches can be combined to learn representations that are not limited to what is observable in the camera~\cite{gasp}.

\parsection{Contrastive learning}
Embedding-space \ac{SSL} methods enforce invariance of representations under semantic-preserving transformations, or predict missing data in embedding space. Preventing trivial solutions is essential in the application of these approaches. Sample-contrastive methods~\cite{simclr, moco} add loss terms that repel representations of different samples and have been applied to point clouds~\cite{masked_scene_contrast}. A newer category of methods is dimension-contrastive and maximizes the information content in batches of embeddings~\cite{barlow_twins, vicreg, sigreg}, with a recent application to LiDAR data~\cite{ad_jepa_wm}.

\parsection{Self-distillation}
In self-distillation~\cite{byol, dino}, architectural asymmetries prevent collapse. The majority of these methods process different views of the input data with separate models. 
One is a model being trained by gradient descent (student) while the other (teacher) typically receives no gradients and is a slow-moving average (usually \ac{EMA}) of the model being directly trained. Current leading \ac{SSL} methods for point clouds are self-distillation-based and trained to produce representations invariant under augmentations~\cite{sonata, bosch_dos}. Recent works~\cite{concerto, utonia} combine self-distillation with distillation of image foundation models. Other works focus on adapting view-invariance \ac{SSL} to become better at distinguishing individual instances of semantic classes~\cite{bosch_pointins, bosch_ghostpoint}. We follow the self-distillation paradigm due to its success in large-scale vision models~\cite{dinov2, dinov3}. 

\parsection{JEPA}
Predictive methods~\cite{a_path_towards} offer an alternative to learning view invariance. In a \ac{JEPA}, one predicts encoded representations of a data sample from representations of a different or incomplete view of that sample. The prediction is carried out by a separate model, the predictor, which is conditioned on latent information about the embedding that has to be predicted (\eg, locations of missing tokens). This paradigm has been applied to images~\cite{ijepa}, and videos~\cite{vjepa, vjepa_2}. Predictive \ac{SSL} has been applied to point clouds in dense \ac{BEV} representations~\cite{ad_l_jepa} and single-object scans~\cite{point_jepa}. To the best of our knowledge, our method is the first \ac{JEPA} applied to sparse transformer architectures on full outdoor point clouds, and we show that it can challenge the best view-invariant methods.

\parsection{Point transformers}
In our choice of model architecture, we take inspiration from the fact that transformers such as ViTs~\cite{vit} surpass domain-specific architectures when trained at scale. Self-attention on every point or fine voxel in a point cloud is intractable in memory and compute. Early point transformers~\cite{ptv1, ptv2} computed separate attention neighborhoods for each voxel and had small receptive fields. PTv3~\cite{ptv3} enabled scaling to larger receptive fields by grouping voxels along space-filling curves, which provide computationally efficient proxies for spatial proximity. 

Despite being named a transformer, more than two-thirds of PTv3's parameters are made up of sparse convolutions~\cite{submanifold_convs}, added to encode relative positions between voxels. LitePT~\cite{litept} extends RoPE~\cite{rope} to 3D positions and enables removal of convolutions from coarse model stages, while relying exclusively on convolutions in fine stages. Prior \ac{SSL} methods like~\cite{utonia} use a similar extension of RoPE, but formulated in a way that is unable to resolve nearby fine voxels. We make adaptations to the use of RoPE that allow us to remove all convolutions, resulting in faster models that are more performant at scale. 

The recent work~\cite{volt} takes an alternative approach to define a pure transformer for point clouds, but we hypothesize that its global attention and coarse tokenization are not scalable to voxel sizes as small, or point cloud ranges as large, as models that use grouping along space-filling curves.

\parsection{Deformable attention}
Our predictor uses a modified version of the deformable attention layer from~\cite{deformable_detr}, designed for efficient information gathering in dense feature maps. We extend the idea to sparse 3D representations, which requires some modifications. 

\parsection{Voxel size generalization}
Point cloud encoders are almost exclusively trained on a single voxel size. Random input scaling during training is common, but only provides robustness to a narrow range of voxel sizes at inference. We adopt ideas from parameterized neural networks in high-energy physics~\cite{param_hep, hep_ewkino}, where unknown variables are treated as additional model inputs and fixed at inference. Inspired by this, we use voxel size as an additional input. Architecturally, the parametrization is performed like text-prompt conditioning in diffusion-based image generators~\cite{dit, adaln_single}.

\section{Loss Scaling}
\label{supmat:loss_scaling}
The loss function is described in \cref{sec:masked_prediction}. This section gives a reminder of the loss function and derives the scaling factor that weights the loss contributions of the predictions at different model stages.
For each stage, $s$, we sample a subset, $\mathcal{Q}_s$, of masked voxel tokens as queries, $q$, at which we predict and supervise. The cardinality of sampled queries $|\mathcal{Q}_s|$ is proportional to the number of occupied voxels in a given sample and stage. To ensure that there is a supervision signal for each sample, we enforce $|\mathcal{Q}_s|\geq 1$ per sample. The loss is defined as
\begin{align}
    \mathcal{L}_{\text{J3PA}} &= \sum_{s=1}^{S} \sqrt{\frac{D_s}{D_5}} \left( \ell_s + \ell'_s \right), \\
    \ell_s &= \frac{1}{|\mathcal{Q}_s| D_s} \sum_{q \in \mathcal{Q}_s} \big\| \hat{f}_{s,q} - \tilde{f}_{s,q} \big\|_1\nonumber\\
    & = \sum_{q \in \mathcal{Q}_s} \ell_{s, q},
\end{align}
where $\tilde{f}_{s,q} = \operatorname{LN}(\operatorname{sg}(f_{s,q})) \in \mathbb{R}^{D_s}$ is the (non-affine) layer-normalized post-U-Net target with stop-gradient, $\hat{f}_{s,q} \in \mathbb{R}^{D_s}$ its prediction, and $\ell_{s, q}$ is the loss term for a single query $q$ at stage $s$. Analogously,  $\ell'_s$ is defined as the loss in stage $s$ for the pre-U-Net target. The following derivation only considers $\ell_{s}$. The equations hold equivalently for $\ell_{s}^{\prime}$, but the pre-U-Net loss gradients do not propagate back from fine to coarse model stages.

The full gradient of the loss at a stage, $\ell_{s}$ with respect to the predictor output at that stage, $\hat{f}_s$, can be written as
\begin{equation}
    \nabla_{\hat{f}_s} \ell_s =
    \begin{bmatrix}
        \left( \nabla_{\hat{f}_{s,1}} \ell_s \right)^\top \\
        \vdots \\
        \left( \nabla_{\hat{f}_{s,|\mathcal{Q}_s|}} \ell_s \right)^\top
    \end{bmatrix}
    \in \mathbb{R}^{|\mathcal{Q}_s| \times D_s}.
\end{equation}
The gradient norm of the loss for stage $s$ is the Frobenius norm of $\nabla_{\hat{f}_s} \ell_s$, which gives
\begin{align}
\big\|  \nabla_{\hat{f}_s} \ell_s \big\|_F^{2} &= \sum_{q \in \mathcal{Q}_s}  \big\|  \nabla_{\hat{f}_{s,q}} \ell_{s, q} \big\|_2^2 \nonumber \\
&= \sum_{q \in \mathcal{Q}_s} \sum_{d=1}^{D_s} \left(\frac{\partial{\ell_{s, q}}}{\partial{\hat{f}_{s, q, d}}}\right)^{2}.
\end{align}
Each partial derivative in the $d$-th entry of the summation over $D_s$ only gets nonzero contributions from the $d$-th channel of $\ell_{s, q}$, and when $\hat{f}_{s,q,d} \neq \tilde{f}_{s,q,d}$, these become
\begin{align}
    \frac{\partial \ell_{s, q}}{\partial \hat{f}_{s,q,d}} \nonumber
    &= \frac{1}{|\mathcal{Q}_s| D_{s}} \operatorname{sign}\!\big(\hat{f}_{s,q,d} - \tilde{f}_{s,q,d}\big)\\
    &= \pm \frac{1}{|\mathcal{Q}_s| D_{s}},
\end{align}
with the sign depending on the prediction being larger or smaller than the target. Summing over all queries and channels gives the squared gradient norm with respect to $\hat{f}_s$,
\begin{align}
    \big\| \nabla_{\hat{f}_s} \ell_s \big\|_F^2 \nonumber
    &= \frac{1}{|\mathcal{Q}_s|^2 D_s^2} \sum_{q \in \mathcal{Q}_s} \sum_{d=1}^{D_s} \left(\pm 1\right)^{2} \\
    &= \frac{1}{|\mathcal{Q}_s| D_s}.
\end{align}
The gradient norm is proportional to $1/\sqrt{D_s}$ for a stage with $D_s$ channels. In the training of each model, all five encoder stages are supervised through their own predictor. The U-Net model structure means that gradients from fine stages propagate through the coarse model stages. Coarse model stages are crucial for capturing semantic information, and the predictive task operating at these stages should contribute at least as much to the model updates as predictive tasks at fine stages. To cancel out the reduced gradient norm caused by the increased channel count in coarse stages, the scale factor $\sqrt{D_s}/\sqrt{D_5}$, is applied when summing the loss terms for the different stages. Note that when distilling a model from a fixed teacher, the channel counts of the teacher model are used to compute the scale factors, because these determine the channel counts of the predictor outputs and the gradient norm discussed above.

The introduced scaling does not completely equalize the expected gradient norm between stages. For each stage, we separately impose an upper limit of $200 \cdot B$ sampled queries for a batch of size $B$. Deeper model stages have fewer occupied voxels, meaning that a smaller fraction of batches reach the imposed limit on the number of queries. The gradient norm of the loss w.r.t. the predictor output is proportional to $1/\sqrt{|\mathcal{Q}_s|}$ as shown above. Hence, stages and samples with fewer supervision queries can get larger gradient norms. This is not compensated for in the loss, as we found empirically that canceling out this effect did not result in model improvements. On the other hand, the $\sqrt{D_s}/\sqrt{D_5}$ scaling empirically led to better models and faster convergence.

\section{Model overview}
\label{supmat:subtle_model_differences}
This section provides additional technical details on the architecture of the J3PA models. First, the encoder architecture and the network capacity of our released models are detailed in \cref{supmat:model_defs}. Second, we give an overview of implementation and architectural differences between our model and PTv3 in \cref{supmat:ptv3_differences}. Finally, additional information about the sparse deformable attention predictor and standard cross-attention variants are provided in \cref{supmat:def_pred}.

\subsection{Encoder architecture} \label{supmat:model_defs}
\parsection{High-level architecture} The J3PA model architecture is based on PTv3~\cite{ptv3}, although without convolutions. The encoder consist of multiple stages of transformer blocks operating on voxel tokens of progressively coarser resolutions. Each stage starts with a pooling layer, collecting voxels from the previous (finer) stage into coarser voxel tokens. Pooling is followed by several attention blocks. In all except the coarsest stage, an unpooling layer combines the features of the given stage with those of a deeper stage, forming a U-Net structure. More attention blocks follow after the unpooling layer. Voxels become larger by a factor two along each edge in transitions from one stage to the next. To avoid costly self-attention between all voxel tokens in a stage, tokens are sorted and grouped by ordering their positions along four space-filling curves, namely two variants of the Hilbert curve~\cite{hilbert_curve} and of the Z-order curve~\cite{zorder}. Tokens within these exclusive groups interact via self-attention, and iterating between different curves makes each voxel get a thorough view of its neighborhood. To make learned weights robust, space-filling curves are randomly permuted during training. This inhibits model weights from depending on the exact neighborhood pattern given by one space-filling curve. For all downstream tasks the layer-normalized post-U-Net outputs of each model stage are used as inputs to the task specific heads.

\begin{figure*}
\includegraphics[width=\linewidth]{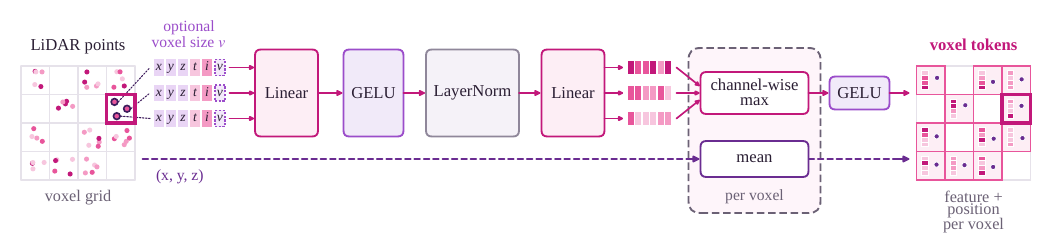}
\caption{Input layer of our point transformer. Each point is represented by its metric $x$, $y$, and $z$ coordinates, time $t$ in seconds, and intensity $i$ mapped to the interval $[0, 255]$. The log-normalized voxel size is an optional additional per-point input, used when training the model to be parametric in the voxel size. All input points are projected by a linear layer followed by GELU activation. A pooling block follows and outputs one feature vector and average point position per voxel.}
\label{fig:initial_embed}
\end{figure*}

\parsection{Initial embedding} The initial voxel tokenization block of our encoder is illustrated in \cref{fig:initial_embed}. Each point in the point cloud is processed by the model, and represented by $x$, $y$, $z$ in metric coordinates, intensities mapped to the range $[0, 255]$, and timestamps in seconds relative to the origin of the input LiDAR scans. The values are projected by a linear layer followed by a GELU activation, and then processed in a way equivalent to the pooling layers between voxel stages. The output of the initial tokenization is a single token and average position per voxel.

\begin{figure}
\includegraphics[width=\linewidth]{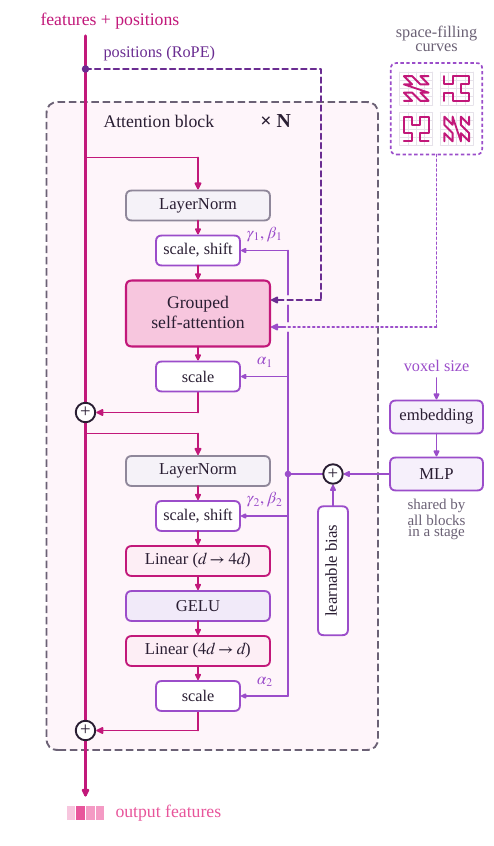}
\caption{Structure of an encoder attention block. Attention and feedforward layers are optionally conditioned on a voxel-size embedding via AdaLN-single conditioning. This is not applied in J3PA-1B which trains on a fixed voxel size.}
\label{fig:encoder_block}
\end{figure}

\parsection{Voxel-size conditioning} The distillation procedure of J3PA-1B into smaller models is parametric in the voxel size, and we use the voxel size as an additional model input. For each sample, a random voxel size is uniformly drawn from an interval $[l, h]$ and used to tokenize the point cloud and determine the voxel sizes at each model stage. The sampled voxel size $v$ is log-normalized as $2 \log_2(v/l)/\log_2(h/l) - 1$ for use as a model input. This value is concatenated to each point as an extra input feature, and additionally converted into an embedding used to condition the attention blocks via AdaLN-single~\cite{adaln_single}. The embedding is created as a Fourier expansion of the log-normalized voxel size, followed by an MLP. An encoder attention block with the AdaLN-single conditioning is shown in \cref{fig:encoder_block}. Using AdaLN-single over standard AdaLN adds far fewer additional parameters to our encoder model, given that every attention block only needs to learn one additional bias vector on top of a shared MLP.

\begin{table*}
\centering
\caption{J3PA model architectures. Per-stage values are listed from the finest to the coarsest stage. The are no post-U-Net layers at the coarsest stage. All values refer to the encoder.}
\label{tab:model_sizes}
\footnotesize
\setlength{\tabcolsep}{3pt}
\begin{tabular}{l r l l l l}
\toprule
Model & Params & Channels & Pre-U-Net layers & Post-U-Net layers & Heads \\
\midrule
Ablation model & 78M  &  64, 128, 256, 512, 1024  & 3, 3, 3, 5, 4 & 1, 1, 1, 1 & 2, 4, 8, 16, 32 \\
J3PA-Small & 42M  & 64, 128, 256, 512, 704     & 2, 2, 2, 2, 4     & 2, 2, 2, 2 & 1, 2, 4, 8, 11 \\
J3PA-Large & 135M & 128, 192, 256, 512, 1024   & 4, 4, 4, 10, 6    & 2, 2, 2, 2 & 2, 3, 4, 8, 16 \\
J3PA-1B    & 1.0B   & 256, 512, 1024, 1536, 1664 & 6, 6, 6, 12, 12   & 2, 2, 2, 2 & 4, 8, 16, 24, 26 \\
\bottomrule
\end{tabular}%
\end{table*}

\parsection{Network capacity} \Cref{tab:model_sizes} defines the number of channels, layers, and attention heads in each of the J3PA models, as well as the total parameter count. The channel widths at every stage of each model are multiples of 64 to make training and inference efficient on modern GPUs. The released models use two attention layers after the U-Net to prevent potential artifacts from having only a single space-filling curve grouping attention after unpooling layers. All models interleave attention operations with standard feedforward networks with a hidden dimension of $4\times$ the channel width and GELU activation functions.

\parsection{Training stability} To train the largest models, J3PA-1B and J3PA-Large, we slightly modify the transformer architecture for stability. The head-specific sigmoid gate from~\cite{sdpa_gate} is applied in all attention blocks. In addition, QK layer normalization~\cite{vit_22b} is used only in the training of J3PA-1B, given that we found self-distillation to be unstable without it at that scale. The released training and inference code provides a custom Triton kernel for QK layer normalization optimized for point transformers, and fused with RoPE. This prevents a large training time and memory overhead.

\parsection{J3PA-131M} The results presented in \cref{tab:less_param_comp} use a 131M parameter model to verify that the J3PA training recipe gives strong results when self-distilled solely on NuScenes data and with fewer model parameters than Utonia. This model is similar to J3PA-Large, but uses fewer channels in the first two stages and has half the channel width per attention head. 

\subsection{Differences from PTv3} \label{supmat:ptv3_differences}
The largest differences between our model and those based on PTv3 are that J3PA does not have any convolution layers and uses the RoPE implementation discussed in \cref{supmat:supp_rope} to retain a strong ability to model positional relationships. On top of this, there are several implementation details and other minor model architecture choices that differ between J3PA and earlier Ptv3-based SSL-models~\cite{sonata, utonia, concerto} that are built in the PointCept~\cite{pointcept} repository.

\parsection{Space-filling curves permutation} Space-filling curves are used to sort and split points into exclusive attention groups. The choice of space-filling curves is randomly permuted as explained in \cref{supmat:model_defs}. In PointCept, the same permutation is applied for an entire batch. Our implementation applies the permutation sample-wise to reduce gradient noise. In addition, when computing voxel grid indices we compute grid minima for every sample separately to make the space-filling curves more aligned to each sample, whereas PointCept uses a batch-wise minimum.

\parsection{Pooling and unpooling layers} As explained earlier, each U-Net stage in both our model and PTv3 has an unpooling layer in which the feature vector of each fine voxel is combined with the feature vector the coarser-stage voxel it falls in. PTv3 applies a linear projection, layer normalization, and activation (in that order) to both the coarse and fine features before summing associated pairs. Our model only applies layer normalization, and a subsequent projection to the coarse voxel features before summing them with the unmodified fine voxel features. This keeps a cleaner residual path throughout the model. Our pooling layers are equivalent to those in PTv3, except for the fact that they are pre-normalized rather than post-normalized, as can be seen in \cref{fig:initial_embed}.

\parsection{Differences in tokenization} To generate the initial embedding in the first model stage, PointCept-based models like PTv3, only retain the information of a single point when multiple points fall inside the same voxel. We opt for retaining all available information and project all points individually. Subsequently we compute channel-wise maximum over the point features in each voxel to get the initial voxel feature. This is illustrated in \cref{fig:initial_embed}.

\subsection{Sparse deformable predictor} \label{supmat:def_pred}
\begin{figure}
\includegraphics[width=\linewidth]{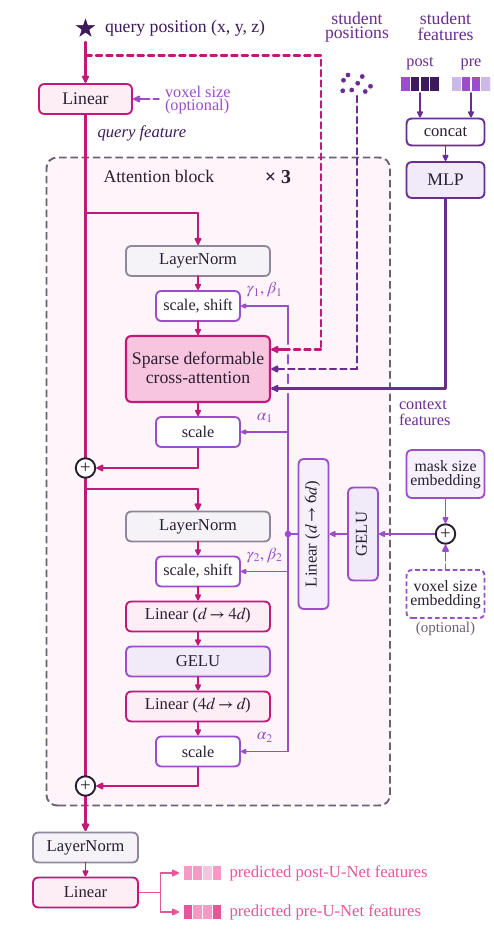}
\caption{Overview of the deformable predictor. Pre- and post-U-Net student features are combined into context features. Given a query feature and location, context information in the query neighborhood is collected and processed by three sparse deformable cross-attention layers. The final output is normalized and projected to twice the channel width of the teacher model and split along the channel axis to make predictions of the pre- and post-U-Net teacher features at the query location. The attention layers are modulated by AdaLN with the mask block size embedding as input, optionally summed with a voxel-size embedding.}
\label{fig:def_predictor_full}
\end{figure}

\parsection{High-level architecture} An overview of the predictor is given in \cref{fig:def_predictor_full}. The predictor consists of several sparse deformable attention blocks built by interleaving deformable attention operations, introduced in \cref{sec:sparse_deformable_predictor}, with query-wise feedforward networks. Each model stage in our \ac{SSL} training has a separate predictor which aims to predict features of the teacher model at masked locations for the corresponding stage. Predictions are based on the available student context features at that model stage.

The student context is constructed by concatenating pre- and post-U-Net features and processing them with a two-layer MLP. Initially, a query consists of a position and a feature vector which is a linear projection of that position. In voxel-size-parametric distillation the log-normalized voxel size (see \cref{supmat:model_defs}) is added as an additional input value when constructing the query feature. Several deformable attention blocks make the query aware of the visible student context. Finally, the query feature is layer normalized and projected to twice the encoder channel width so it can be split along the channel dimension into a pre- and post-U-Net teacher feature prediction.

\parsection{Predictor sizes} Each model training presented in this work uses predictors consisting of 3 attention blocks and an internal channel width equal to that of the stage of the encoder model providing the student context (see \cref{tab:predictor_sizes}). Compressing the internal channel width of the predictor relative to the encoder, as is common in \ac{JEPA} for videos~\cite{vjepa, vjepa_2, vjepa_2p1}, did not lead to improved performance in small-scale trainings of the ablation model. Similarly, adding additional predictor layers did not result in increased model performance. Due to excessive compute requirements, we have not tested multiple configurations in large scale model trainings like J3PA-1B.

\begin{table*}
\centering
\caption{Predictor architectures for each model training. The predictor is independent for each model stage and values are listed for all predictors from the finest to the coarsest stage. The number of layers and head-wise channel widths are the same for every stage. We present the internal channel counts used by the predictor, but note that output features are always projected to match teacher channel counts.}
\label{tab:predictor_sizes}
\footnotesize
\setlength{\tabcolsep}{3pt}
\begin{tabular}{l l l l l}
\toprule
Enc. Model & Params (M) & Channels & Num. layers & Head channels \\
\midrule
Ablation model & 0.29, 1.16, 4.61, 18.4, 73.6   & 64, 128, 256, 512, 1024    & 3 & 32 \\
J3PA-Small     & 0.25, 0.98, 3.87, 14.9, 27.5   & 64, 128, 256, 512, 704     & 3 & 64 \\
J3PA-Large     & 0.91, 2.09, 3.87, 14.9, 56.6   & 128, 192, 256, 512, 1024   & 3 & 64 \\
J3PA-1B        & 4.58, 18.3, 73.0, 164.2, 192.7 & 256, 512, 1024, 1536, 1664 & 3 & 64 \\
\bottomrule
\end{tabular}
\end{table*}

\parsection{Conditioning predictor} The predictor is always conditioned on the size of the mask cuboids, which consist of three randomly drawn edge lengths per training sample. Conditioning is done via AdaLN~\cite{dit} in the attention and FFN blocks (J3PA-Small, J3PA-Large) or by applying FiLM~\cite{film} to the outputs of attention and FFN layers (J3PA-1B). Fig.~\ref{fig:def_predictor_full} shows the AdaLN variant of the predictor. The input to the conditioning mechanism is a mask size embedding with the same channel width used by the predictor layers. Mask size embeddings are extracted by reading out separate learned embedding tables for each of the cuboid size axes $x$, $y$, and $z$ and summing the results. In our voxel-size-parametric distillation procedure the predictor blocks are additionally conditioned on a voxel-size embedding (constructed equivalently as for encoder conditioning), which is summed with the mask embedding and projected into one conditioning embedding.

\subsection{Standard cross-attention}
\Cref{sec:ablate_predictor} ablated several predictor designs and compared our sparse deformable attention predictor against standard cross-attention. For completeness, we describe how the cross-attention predictor was implemented. In the most naive setup, each query would attend all context tokens. This, however, leads to an excessive compute requirement for dense point clouds. A solution is to compute a neighborhood of $K$ context tokens that each query should attend. Using an exact Euclidean nearest neighbor search can retrieve the relevant context tokens, but becomes inefficient quickly as $K$ increases.

A faster alternative is to use space-filling curves, as done when grouping tokens in the self-attention operations of the encoder model. The optimal neighborhood for a given query and space-filling curve is found as its nearest neighbors along that curve. This can be computed quickly for large neighborhoods compared to a Euclidean distance-based search. By using different space-filling curves in subsequent cross-attention layers, each query can thoroughly explore its neighborhood. This is the implementation used in the aforementioned ablation study, and the context neighborhood for each query along each space-filling curve consists of 256 tokens.

The design described above leads to training iterations that are memory intensive and slow since each query has to explore a separate neighborhood. A more efficient implementation is to group the set of context tokens into exclusive groups along the space-filling curves, and then to use one of the fixed context groups for cross-attention with a given query. The downside is that this leads to a coarser context neighborhood assignment for the queries.

\section{RoPE} \label{supmat:supp_rope}
As described in the main body of the paper we make modifications in the application of RoPE~\cite{rope} compared to earlier work on \ac{SSL} for point clouds.

\parsection{Resolving nearby voxels} First, we identify a limitation in RoPE's application in Utonia~\cite{utonia}. To train a single encoder for all point clouds, Utonia rescales outdoor scenes by a factor $1/5$ and uses a voxel size of $0.01$ in the rescaled space, corresponding to $5$ cm in metric units. RoPE is applied to the rescaled point coordinates, yielding positional encodings in which the shortest wavelength is $\pi \cdot 10$ m $\approx 31 $ m, or 628 voxel edges in the finest model stage. As shown in Sec~\ref{sec:ablate_rope}, using this RoPE configuration without convolutional layers in the model results in poor performance. We hypothesize that this is caused by only a weak signal being available to distinguish nearby voxels at fine model stages. In Utonia a frequency base of $10$ is used instead of $10000$ as in the original RoPE design, which leads to larger angles and hence more focus on short positional differences. However, this does not make much difference in practice as all three wavelengths in their attention heads remain very long compared to the size of voxels.

We address this by providing the input positions to RoPE in units of voxel edge lengths for each stage. This results in the shortest positional encoding wavelength being only $2\pi$ voxel edges for each model stage, enabling the model to easily distinguish nearby voxels by position. We use a frequency base of 100, given that overly small values would lead to positional encodings that lack long-range dependencies. Using voxel-edge units as RoPE input positions is similar to how it is applied in the deeper stages of LitePT~\cite{litept}.

\parsection{Interleaving coordinate axes and base angles} In its original 1D formulation, RoPE uses $D/2$ angles to generate positional embeddings for $D$-channel feature vectors~\cite{rope}. Each angle is used to make a $2\times2$ rotation matrix that rotates two adjacent feature channels. When extending RoPE to 3D positions we assign an equal number of channels to the positional encoding of each of the coordinate axes, resulting in $D/6$ angles. One can either enforce the model channel widths to be multiples of $6$ to guarantee a consistent amount of RoPE angles for each coordinate axis, or one can make $\lceil{D/6\rceil}$ angles for any even channel width $D$ and truncate it to $D/2$ angles. Utonia and LitePT opt for the former approach and apply the same RoPE rotations to each attention head separately, so that their total channel counts, and even the channel counts of the individual heads, must be multiples of 6. This is inefficient on modern GPUs which are optimized for channel widths that are multiples of 32 or 64, hence we take the latter approach and truncate the set of angles.

We build the RoPE rotation matrix from the entire channel width $D$ of the queries and keys, before they are split into different attention heads. In the 3D extension of RoPE we consider, each $2\times2$ rotation matrix is sensitive to only one of the coordinate axes and corresponds to a fixed wavelength. It is desirable for all attention heads to be sensitive to position differences along all three coordinate axes, and each head should therefore contain channels rotated by angles depending on the different axes. Additionally, each attention block should a priori have the information necessary to model both short- and long-range positional dependencies. If one simply stacks $D/6$ rotation matrices per axis on top of each other, most attention heads will only be sensitive to one of the coordinate axes and have only low- or high-frequency positional embeddings. A careful interleaving of the different angles is needed to make the rotation matrix handle this, for which pytorch code is provided in \cref{code:rope}. Note that during model training one would precompute the indices used to interleave the angles. An advantage of our scheme compared to using the same angles in each attention head, is that we retain a richer positional embedding base and minimize the impact of truncating the set of angles when the model channel width is not a multiple of 6.

\begin{figure*}[t]
\begin{lstlisting}[
  style=pytorch,
  caption={Pytorch example of the RoPE axis and angle interleaving scheme used in the training of J3PA. This interleaving scheme assumes the multi-head attention implementation splits the query and key vectors along the channel axis into different heads. The indices can be precomputed to speed up training.},
  label={code:rope}
]
def apply_rope(feat, pos, num_heads):
    dim = feat.shape[-1]
    pos_dim = pos.shape[-1]
    dh = dim // num_heads
    angles_per_axis = int(math.ceil(dh/(2*pos_dim))*num_heads)
    angles = max_freq ** (-2 * torch.arange(angles_per_axis) / (dim / pos_dim))

    head_idx = torch.arange(dim) // dh
    pair_idx = (torch.arange(dim) %
    angle_idx = (pair_idx //pos_dim) * num_heads + head_idx
    pos_idx = (pair_idx + head_idx) %

    angles = angles[angle_idx].unsqueeze(0) * pos[..., pos_idx]
    cos = angles.cos()
    sin = angles.sin()
    interleaved_feat = torch.stack([-feat[..., 1::2], feat[..., ::2]], dim=-1).flatten(-2)
    return cos * feat + sin * interleaved_feat

\end{lstlisting}
\end{figure*}

\section{Training Details} \label{supmat:training-details}
We present details and hyperparameters of our training recipe in \cref{supmat:hyperparams}, and describe the training data corpus in \cref{supmat:training_data}.

\subsection{Training hyperparameters} \label{supmat:hyperparams}
Important hyperparameters for our \ac{SSL} model training runs are given in \cref{tab:hyperparams} for reproducibility. For a complete view of all details necessary to train J3PA as presented here we refer to our released training code. All models are trained with the AdamW~\cite{adamw} optimizer, a linear learning rate warm-up, and a subsequent cosine annealing of the learning rate. 
We disable weight decay on all bias terms, norm affines, conditioning layers, offset prediction layers in deformable attention, and embedding tables. The teacher EMA increases along a linear schedule over the course of training. The mask cuboid side lengths are sampled independently for each stage and axis, as integers in units of voxels per axis.

The total training schedule, which determines the cosine annealing of the learning rate and the linear increase of the EMA, stretches over 195 000 training steps in the training of the main models (shorter for the ablation models as specified in \cref{sec:ablations}). In practice, model training runs were truncated early due to time constraints. J3PA-1B trained for approximately 100 000 steps, meaning it did not reach its peak EMA value or the end of the cosine learning rate annealing. Downstream task results kept improving with training length up to the point of truncation, meaning that further scaling should be an avenue towards future performance gains.

\begin{table*}
\centering
\caption{Hyperparameters used in the training of our released \ac{SSL} models.}
\label{tab:hyperparams}
\footnotesize
\begin{tabular*}{\textwidth}{@{\extracolsep{\fill}} l c c}
\toprule
Parameter & J3PA-1B & J3PA-Large/Small \\
\midrule
Number of aggregated scans & $\mathcal{U}(1, 3)$ & $\mathcal{U}(1, 3)$ \\
Crop size & $50\times50$ $\text{m}^{2}$ & $50\times50$ $\text{m}^{2}$ \\
Context window fraction & $\sim \mathcal{U}(0.8, 1.0)$ per axis & $\sim \mathcal{U}(0.8, 1.0)$ per axis \\
Voxel edge size & 5 cm & $\sim \mathcal{U}(4, 20)$ cm \\
Expected queries per batch & $200 \times B$ & $200 \times B$ \\
Def. Attn. offsets per head & 3 & 3 \\
Def. Attn. neighbors per offset & 2 & 2 \\
Masking fraction (per stage) & 0.15, 0.15, 0.3, 0.3, 0.3 & 0.15, 0.15, 0.3, 0.3, 0.3 \\
Mask cuboid edge size (random per axis) & $\mathcal{U}(4, 10)$ & $\mathcal{U}(4, 10)$ \\
Warmup steps & 1900 & 1000 \\
Global batch size ($B$) & 128 & 128 \\
Starting learning rate & $10^{-8}$ & $10^{-8}$ \\
Peak learning rate & $5 \times 10^{-4}$ & $5 \times 10^{-4}$ \\
Final learning rate (not reached) & $2.5 \times 10^{-5}$ & $2.5 \times 10^{-5}$ \\
EMA start & 0.999 & 0.999 \\
EMA end (not reached) & 0.9995 & 0.9995 \\
Weight decay & 0.05 & 0.05\\
\bottomrule
\end{tabular*}
\end{table*}

\subsection{Training Data}
\label{supmat:training_data}
\begin{table}
\centering
\caption{Training corpus for J3PA. We only use the training splits during \ac{SSL} training, except for the $20$k unlabeled AV2 LiDAR scenes~\cite{argoverse2}. Validation splits are used to evaluate downstream probes, and the test splits are never used.}
\label{tab:data-splits}
\footnotesize
\setlength{\tabcolsep}{3pt}
\begin{tabular*}{\columnwidth}{@{\extracolsep{\fill}}l rr rr r@{}}
\toprule
& \multicolumn{2}{c}{\textbf{Train split}} & \multicolumn{2}{c}{\textbf{Val split}} & \multicolumn{1}{c}{\textbf{Test split}} \\
\cmidrule(r){2-3} \cmidrule(lr){4-5} \cmidrule(l){6-6}
& \multicolumn{2}{c}{Pre-training} & \multicolumn{2}{c}{Evaluation} & \multicolumn{1}{c}{Never used} \\
\cmidrule(r){2-3} \cmidrule(lr){4-5} \cmidrule(l){6-6}
Dataset & Seq. & Sweeps & Seq. & Frames & Seq./sweeps \\
\midrule
ZOD~\cite{zod} & 90.0k & 1.89M  & 10.0k & 10.0k & -- \\
Argoverse 2~\cite{argoverse2}  & 20.7k & 6.09M  & 150   & 23.5k & 150/23.6k \\
nuScenes~\cite{nuscenes}     & 700   & 273k   & 150   & 6.0k  & 150/58.4k \\
TruckScenes~\cite{truckscenes}  & 523   & 105k   & 75    & 3.0k  & 149/29.9k \\
OctoSense~\cite{octosense}    & 293   & 1.68M  & --    & --    & 78/440k \\
PandaSet~\cite{pandaset}     & 72    & 5.8k   & 31    & 2.5k  & -- \\
KITTI~\cite{kitti_raw}        & 106   & 37.2k  & 45    & 3.8k  & --/7.5k \\
Cirrus~\cite{cirrus}       & 6     & 10.7k  & 1     & 897   & -- \\
\midrule
J3PA corpus & 112k  & 10.1M & -- & -- & --\\
\bottomrule
\end{tabular*}
\end{table}

Tab. \ref{tab:data-splits} gives an overview of the data corpus used in the training of J3PA. We do not use the validation or test splits for any dataset, with the notable exception of the AV2 LiDAR \ac{SSL} Dataset~\cite{argoverse2}, where we use all splits. The reason is that the validation and test splits of this dataset have no labels to evaluate on. Avoiding the use of validation and test splits ensures that our models can be used as unbiased backbones in the evaluation of downstream methods on common benchmarks. With the argument that there are no labels to overfit on during \ac{SSL} training, other commonly available LiDAR \ac{SSL} models~\cite{sonata, utonia, concerto}, have opted to use test and validation sets of common benchmark datasets during training. We found it difficult to reason about the potential impact this has on benchmarks and have decided to avoid this possible source of bias.

\section{Additional Results}
\label{supmat:additional-results}
This section covers the evaluation protocol used in the main paper in \cref{supmat:eval}, and a simpler evaluation pipeline representative of inference time performance in \cref{supmat:single_forward}. Results on a collection of datasets when using the simplified evaluation protocol are presented in \cref{supmat:other_dataset}.

\subsection{Evalution protocol}
\label{supmat:eval}
We evaluate the representation quality of our models by freezing the encoders and training probes that use their output features. LiDAR point features are obtained by concatenating the post-U-Net outputs of all attention stages for the voxels containing a given point. For semantic segmentation the probe consists of a single linear layer, whereas in panoptic segmentation an additional 2 layer MLP for offset-based instance determination is trained alongside the single-layer semantic head as is done for panoptic linear probe training in PointGroup~\cite{pointgroup}.

The evaluation protocol follows the PointCept repository~\cite{pointcept}. For each point cloud in the evaluation set, the sample is augmented with several \acp{TTA}. For classification, the output probabilities are summed for all model outputs over the \acp{TTA} of the given point cloud. The class with the highest summed value per point determines the final classification decision for that point. For offset predictions, each prediction is converted back into the non-augmented view of the point cloud. Predictions are then averaged to produce the final offset. In PointCept-based models like Utonia, the number of \acp{TTA} varies depending on the number of LiDAR points falling in the initial voxels. The total number of forward passes used to evaluate these methods is typically over 100. Our protocol limits the number of forward passes to exactly 96 for each point cloud. We use the same augmentations as PointCept with the exception of the voxel tokenization, where they choose one point per voxel and vary which point to pick. Our model encodes all points making this augmentation not as well-posed. Instead, we slightly shift the grid to achieve a similar effect of small variations in the initial voxel tokens. 

\subsection{Single forward pass}
\label{supmat:single_forward}
However effective for offline settings without compute constraints, running $\sim 100$ forward passes per LiDAR sweep is intractable on embodied platforms or systems with real-time requirements. We believe that a simpler evaluation procedure would give a more accurate summary of the real world performance of trained models when deployed. Therefore, we provide an alternative evaluation pipeline where model predictions are obtained from a single forward pass on a non-augmented data sample. In all other aspects, this evaluation protocol still follows that in PointCept. Results under the simplified evaluation protocol are provided in \cref{tab:single-forward-eval}. 

\subsection{More dataset results.}
\label{supmat:other_dataset}
We report linear probe performance for a single forward pass per sample on several datasets in \cref{tab:single-forward-eval}. One should note that Utonia~\cite{utonia} is evaluated out-of-distribution for all datasets except nuScenes~\cite{nuscenes}, Waymo~\cite{waymo} and SemanticKITTI~\cite{semantic_kitti} which are used during its training. This is comparable to our model being out-of-distribution on the Waymo dataset.

For datasets without point-wise semantic or panoptic labels, but containing object bounding box annotations, we make point-wise ground truth labels by marking all points falling inside a bounding box with the class label of the box. All points falling outside of a bounding box are treated as a single background class. 

\newcommand{\dsrot}[1]{\rotatebox[origin=c]{90}{\scriptsize\shortstack{#1}}}
\begin{table}[t]
\centering
\caption{Linear probe performance evaluated with a single forward pass per point cloud. Best result per dataset in \textbf{bold}.}
\label{tab:single-forward-eval}
\footnotesize
\setlength{\tabcolsep}{4pt}
\begin{tabular}{@{}c@{\hspace{4pt}}lrrrrrr@{}}
\toprule
& Methods & mIoU & mAcc & allAcc & PQ & RQ & SQ \\
\midrule
\multirow{4}{*}{\dsrot{nuScenes}}
& Utonia (lin.) & 70.4 & 84.1 & 93.1 & 65.7 & 76.4 & 85.2 \\
& \cellcolor{jPinkBg} J3PA-Small (lin.) & 72.4 & 84.6 & 94.3 & 70.9 & 80.8 & 86.9 \\
& \cellcolor{jPinkBg} J3PA-Large (lin.) & 76.5 & 86.9 & 94.8 & 75.0 & 84.7 & 88.0 \\
& \cellcolor{jPinkBg} J3PA-1B (lin.) & \textbf{81.4} & \textbf{89.4} & \textbf{95.3} & \textbf{78.9} & \textbf{88.1} & \textbf{89.2} \\
\midrule
\multirow{4}{*}{\dsrot{Semantic\\KITTI}}
& Utonia (lin.) & 67.1 & 75.1 & 92.2 & 56.0 & 65.8 & 78.1 \\
& \cellcolor{jPinkBg} J3PA-Small (lin.) & 66.7 & 76.3 & 92.4 & 54.7 & 64.8 & 77.6 \\
& \cellcolor{jPinkBg} J3PA-Large (lin.) & 68.8 & 76.8 & 92.5 & 57.6 & 67.7 & \textbf{78.6} \\
& \cellcolor{jPinkBg} J3PA-1B (lin.) & \textbf{71.0} & \textbf{77.7} & \textbf{93.1} & \textbf{60.6} & \textbf{70.6} & 75.9 \\
\midrule
\multirow{4}{*}{\dsrot{Waymo}}
& Utonia (lin.) & 65.3 & 77.6 & 93.4 & 51.6 & 62.4 & 80.3 \\
& \cellcolor{jPinkBg} J3PA-Small (lin.) & 64.2 & 76.9 & 93.5 & 53.4 & 65.2 & 80.0 \\
& \cellcolor{jPinkBg} J3PA-Large (lin.) & 66.4 & 77.7 & 94.0 & 56.6 & 68.2 & 81.0 \\
& \cellcolor{jPinkBg} J3PA-1B (lin.) & \textbf{70.0} & \textbf{81.2} & \textbf{94.5} & \textbf{61.4} & \textbf{73.3} & \textbf{82.6} \\
\midrule
\multirow{4}{*}{\dsrot{Argo-\\verse 2}}
& Utonia (lin.) & 27.9 & 47.9 & 80.6 & 27.1 & 32.4 & 64.3 \\
& \cellcolor{jPinkBg} J3PA-Small (lin.) & 34.2 & 52.5 & 82.2 & 33.9 & 40.5 & 66.4 \\
& \cellcolor{jPinkBg} J3PA-Large (lin.) & 39.1 & 56.3 & 84.9 & 37.7 & 44.8 & \textbf{68.9} \\
& \cellcolor{jPinkBg} J3PA-1B (lin.) & \textbf{42.9} & \textbf{58.6} & \textbf{85.8} & \textbf{41.1} & \textbf{48.7} & 68.0 \\
\midrule
\multirow{4}{*}{\dsrot{ZOD}}
& Utonia (lin.) & 42.7 & 61.2 & 85.0 & 30.5 & 37.6 & 75.0 \\
& \cellcolor{jPinkBg} J3PA-Small (lin.) & 48.7 & 68.4 & 88.5 & 39.6 & 48.0 & 77.3 \\
& \cellcolor{jPinkBg} J3PA-Large (lin.) & 51.6 & 71.4 & 89.5 & 42.6 & 51.3 & 77.7 \\
& \cellcolor{jPinkBg} J3PA-1B (lin.) & \textbf{55.1} & \textbf{74.9} & \textbf{90.3} & \textbf{47.1} & \textbf{56.1} & \textbf{79.0} \\
\midrule
\multirow{4}{*}{\dsrot{PandaSet}}
& Utonia (lin.) & 32.3 & 42.1 & 90.0 & 24.4 & 29.9 & 61.7 \\
& \cellcolor{jPinkBg} J3PA-Small (lin.) & 36.9 & 44.3 & 92.7 & 29.3 & 35.8 & 61.3 \\
& \cellcolor{jPinkBg} J3PA-Large (lin.) & 37.8 & 45.0 & 93.1 & 30.8 & 37.4 & 62.3 \\
& \cellcolor{jPinkBg} J3PA-1B (lin.) & \textbf{41.9} & \textbf{49.9} & \textbf{93.8} & \textbf{35.1} & \textbf{42.5} & \textbf{65.0} \\
\midrule
\multirow{4}{*}{\dsrot{Truck-\\Scenes}}
& Utonia (lin.) & 51.5 & 76.0 & 83.8 & 34.8 & 42.6 & 81.5 \\
& \cellcolor{jPinkBg} J3PA-Small (lin.) & 57.2 & 76.1 & 88.1 & 46.2 & 56.4 & 79.7 \\
& \cellcolor{jPinkBg} J3PA-Large (lin.) & 60.0 & 77.1 & 89.2 & 50.5 & 61.6 & 79.7 \\
& \cellcolor{jPinkBg} J3PA-1B (lin.) & \textbf{66.2} & \textbf{83.8} & \textbf{89.8} & \textbf{57.2} & \textbf{68.4} & \textbf{83.1} \\
\midrule
\multirow{4}{*}{\dsrot{Cirrus}}
& Utonia (lin.) & 48.3 & 61.9 & 96.4 & 27.1 & 32.3 & \textbf{54.2} \\
& \cellcolor{jPinkBg} J3PA-Small (lin.) & 47.9 & 59.6 & \textbf{97.3} & 27.3 & 32.1 & 53.4 \\
& \cellcolor{jPinkBg} J3PA-Large (lin.) & \textbf{54.2} & \textbf{68.0} & \textbf{97.3} & 29.4 & 34.7 & 53.1 \\
& \cellcolor{jPinkBg} J3PA-1B (lin.) & 53.3 & 63.1 & \textbf{97.3} & \textbf{29.7} & \textbf{35.8} & 51.7 \\
\bottomrule
\end{tabular}
\end{table}

\subsection{Qualitative}
\label{supmat:qualitative}
For all \ac{PCA} visualizations we calculate the principal axes from 24 object-rich ZOD validation frames and select $3$ components among the top $7$ with the most variance. We also vary the order of the RGB values to maximally show the informational content of the learned features. 

\begin{figure*}[t]
  \centering
  \pcaimg{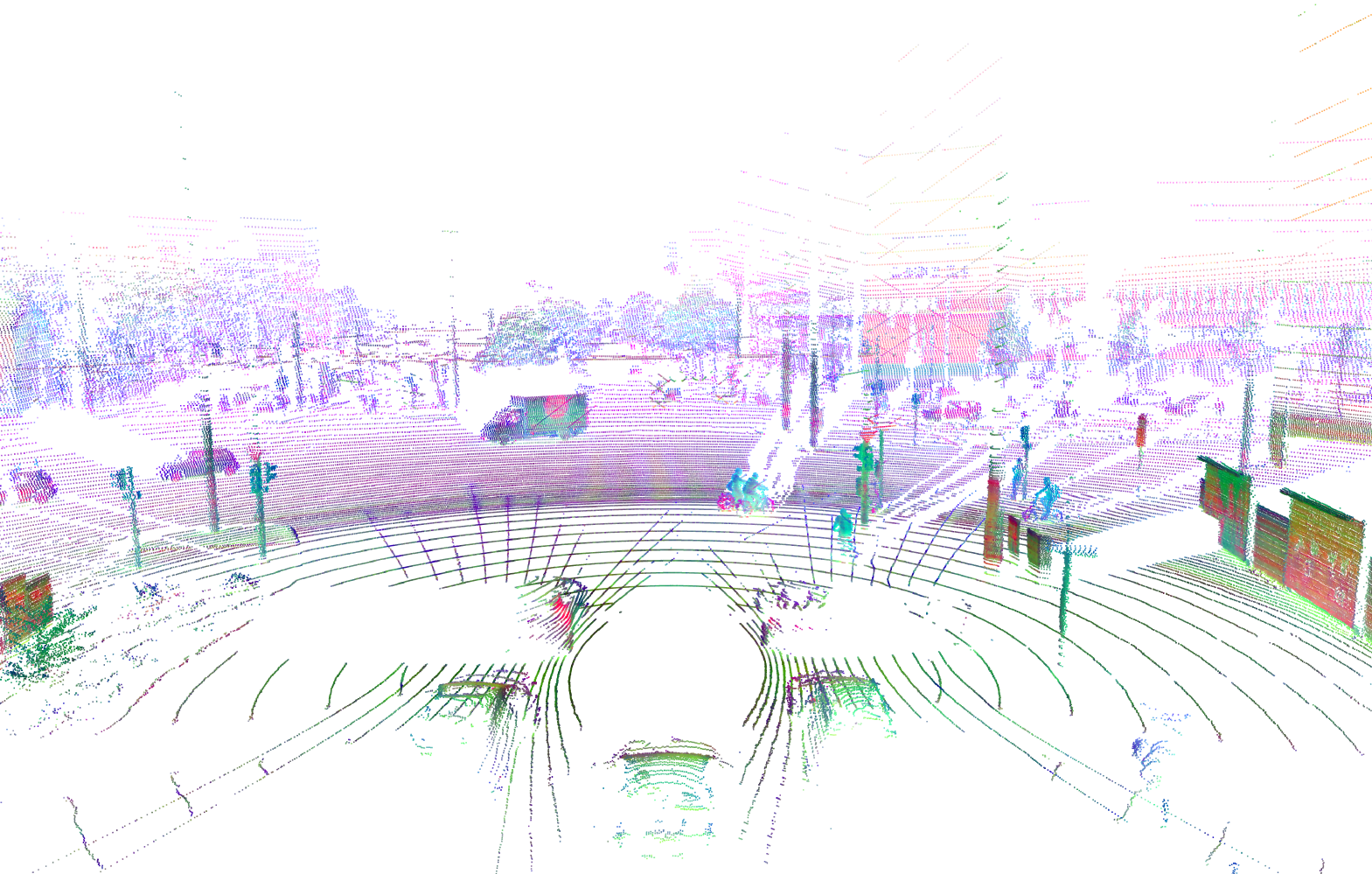}
  \caption{\ac{PCA} visualization of J3PA-1B features at stage 1. }
  \label{fig:pca_stages:s1}
\end{figure*}

\begin{figure*}[t]
  \centering
  \pcaimg{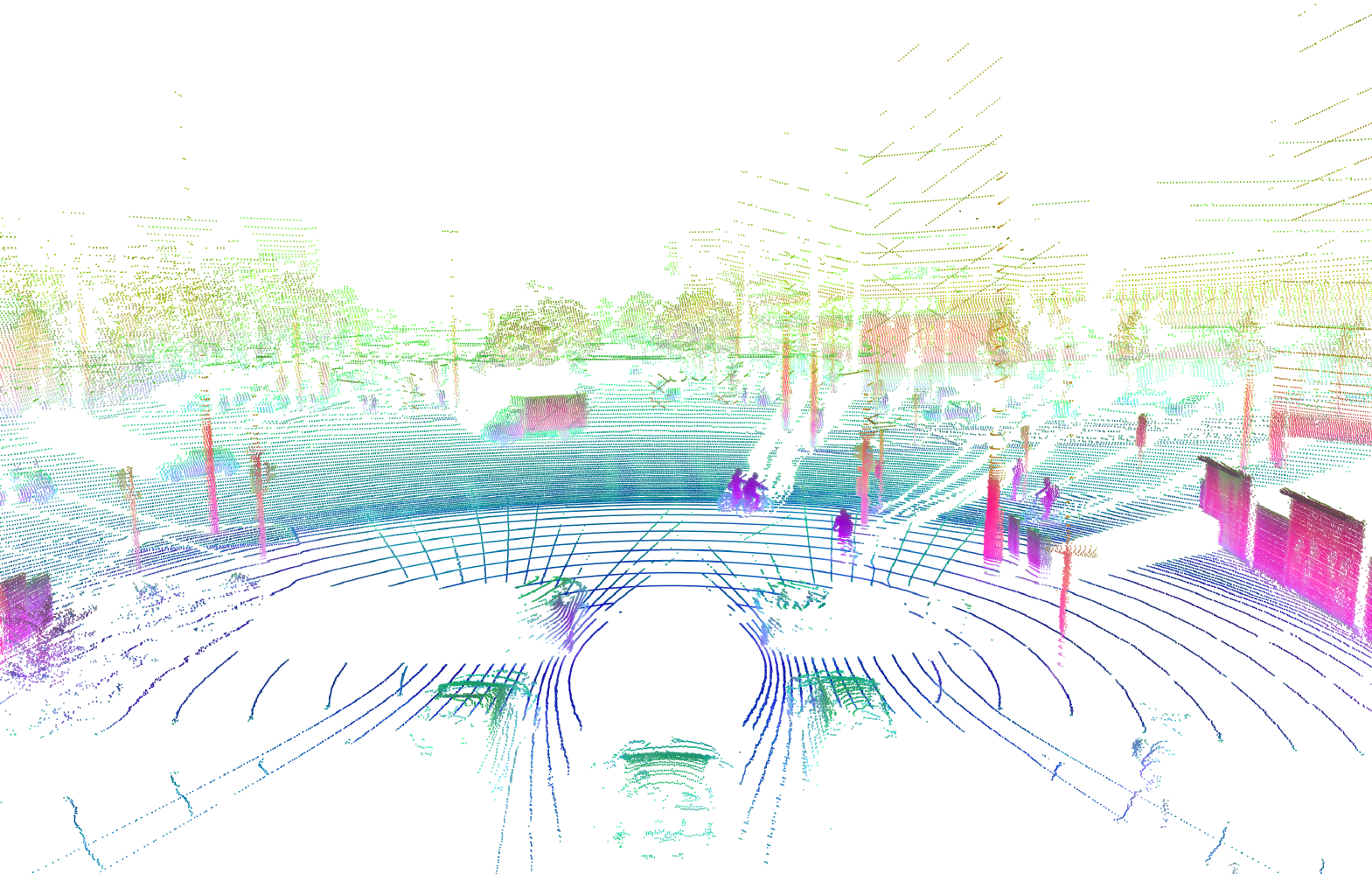}
  \caption{\ac{PCA} visualization of J3PA-1B features at stage 2.}
  \label{fig:pca_stages:s2}
\end{figure*}

\begin{figure*}[t]
  \centering
  \pcaimg{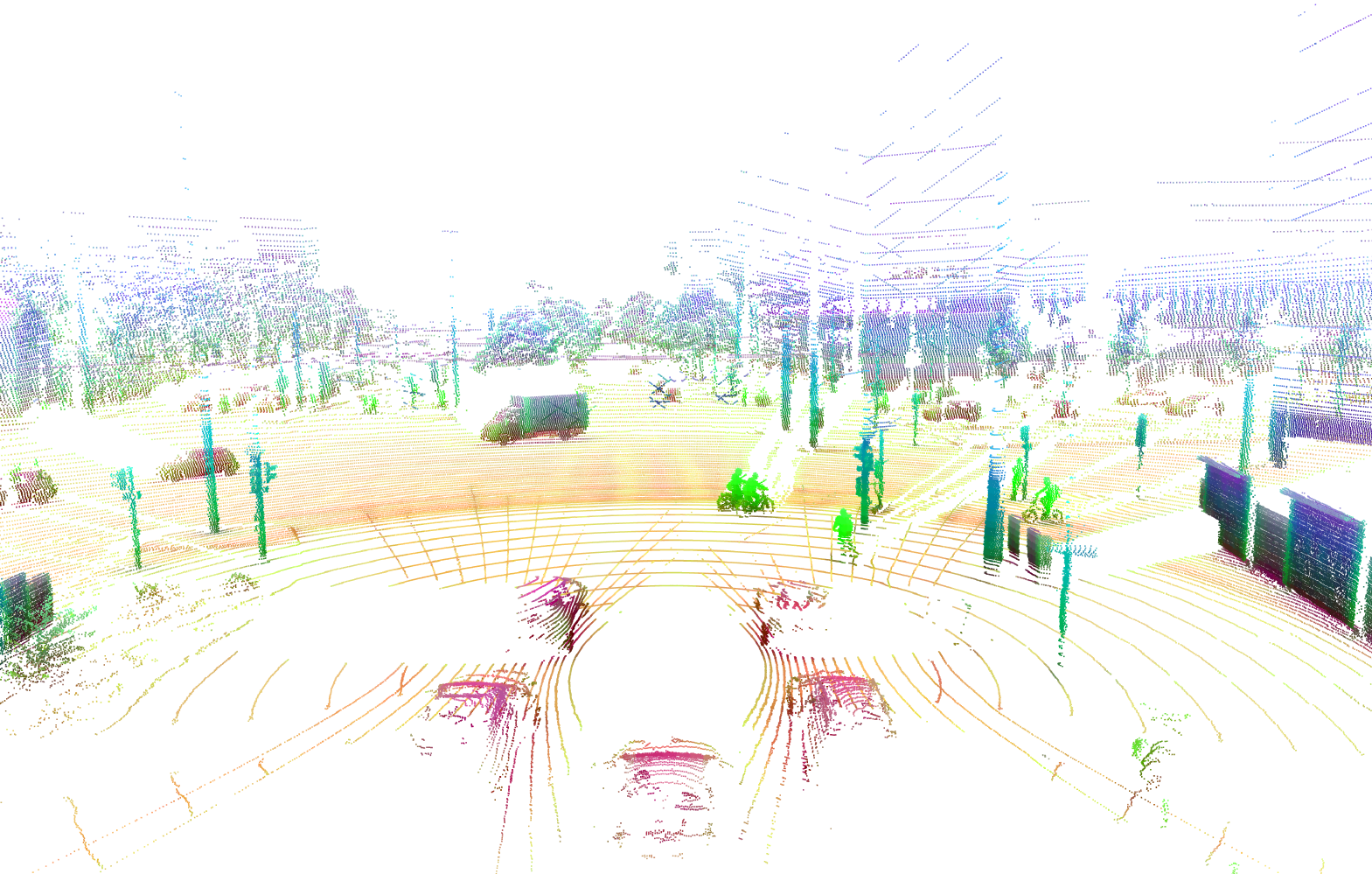}
  \caption{\ac{PCA} visualization of J3PA-1B features at stage 3.}
  \label{fig:pca_stages:s3}
\end{figure*}

\begin{figure*}[t]
  \centering
  \pcaimg{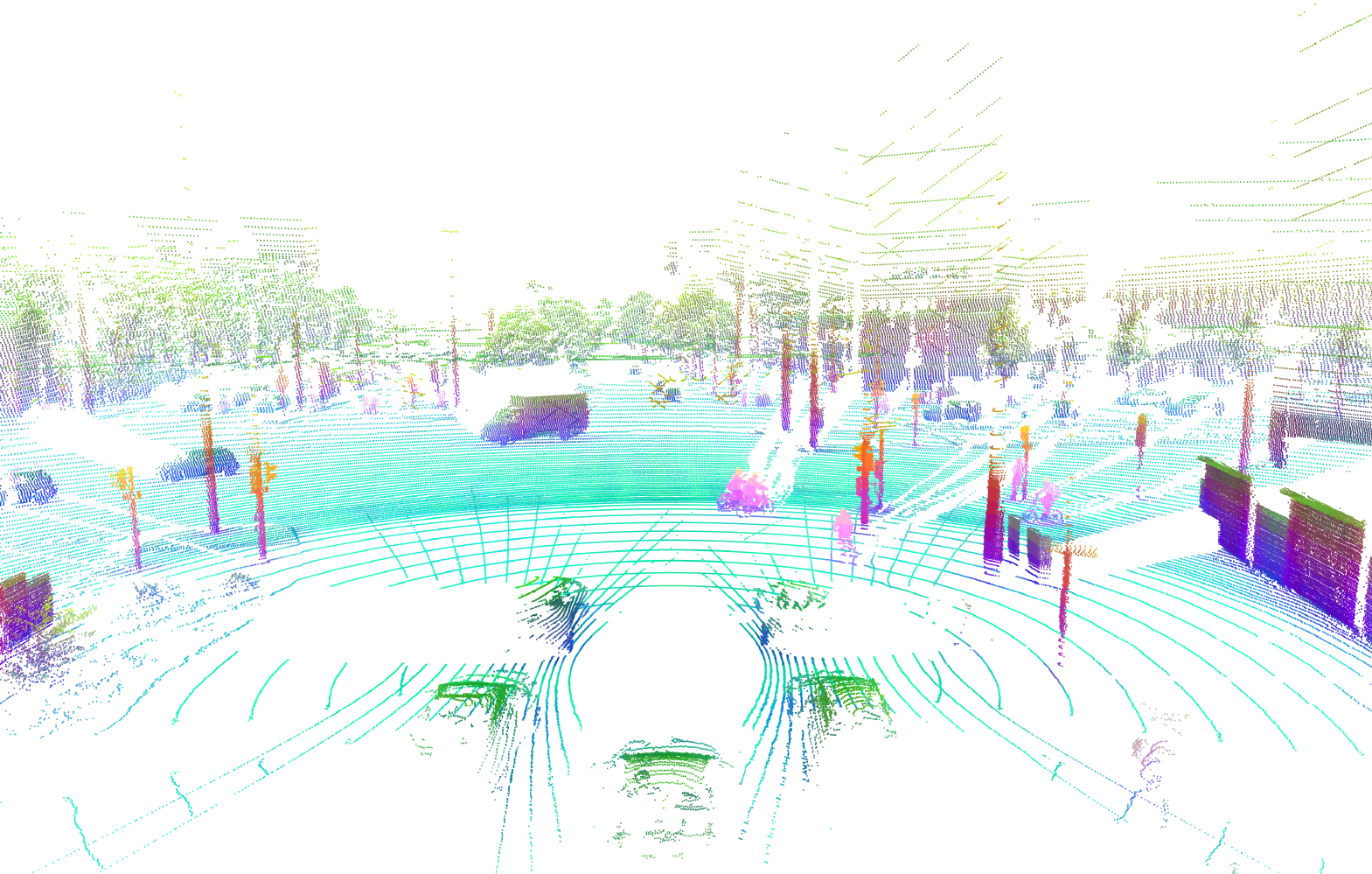}
  \caption{\ac{PCA} visualization of J3PA-1B features at stage 4.}
  \label{fig:pca_stages:s4}
\end{figure*}

\begin{figure*}[t]
  \centering
  \pcaimg{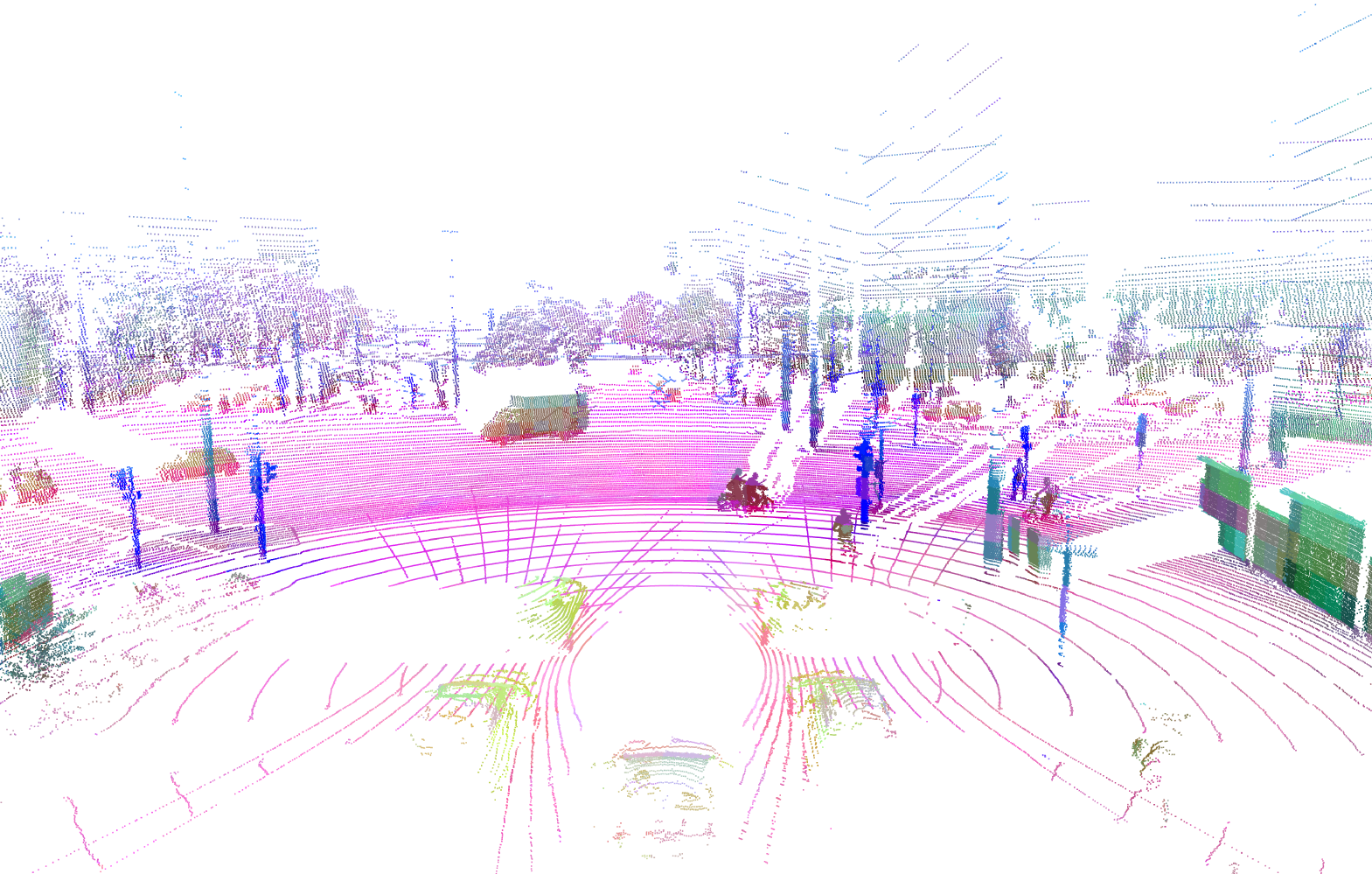}
  \caption{\ac{PCA} visualization of J3PA-1B features at stage 5.}
  \label{fig:pca_stages:s5}
\end{figure*}

In \cref{fig:pca_stages:s1,fig:pca_stages:s2,fig:pca_stages:s3,fig:pca_stages:s4,fig:pca_stages:s5} it is noticeable that earlier stages contain more fine-grained information, like separating a bicycle from the bicyclist, and the wheels from a car. Later stages merge these objects and can instead identify larger objects such as a car. In the deepest stage, artifacts from voxels containing multiple semantic classes can be seen, and object parts are no longer well-separated. This is in line with the large primary masks at this stage. The voxel features are assigned to all points contained in that voxel voxel for denser visualizations.

We further supply supplementary videos of stage 3 PCA visualizations on unseen ZOD sequences~\cite{zod}, illustrating consistency and robustness of the J3PA-1B representations.

\paragraph{Video 1: Stage 3 — Aggregated PCA visualization showing the consistency of J3PA-1B embeddings}
(\texttt{video1\_aggregated\_pca.mp4}, 18\,s).
The ZOD sequence 000168 is encoded per-sweep with J3PA-1B and the \ac{PCA} colors are extracted for each sweep and accumulated and plotted in a common world frame as the vehicle drives. Despite each sweep being encoded separately, the embeddings show remarkable consistency as concepts and instances keep their color over time.

\paragraph{Video 2: Stage 3 — Front view PCA visualization showcasing the robustness of J3PA-1B embeddings}
(\texttt{video2\_front\_view\_pca.mp4}, 18\,s).
This video shows the per-sweep \ac{PCA} visualization on ZOD sequence 001114 from a chase camera point of view that follows the vehicle through the sequence. The front camera video is temporally aligned and shown top left for reference only and is not used by the model. Objects such as pedestrians, cars, and even flags stay clearly consistent despite occlusions, crowds, and a changing viewpoint.


\end{document}